\RequirePackage{fix-cm}
\documentclass[twocolumn]{svjour3}          
\smartqed  
\usepackage{graphicx}
\usepackage{subfigure}
\usepackage{amsmath}
\usepackage{color}
\usepackage{algorithm}
\usepackage{algpseudocode}
\usepackage{booktabs}
\usepackage{natbib}
\usepackage{hyperref}
\hypersetup{colorlinks=true,citecolor=blue,linkcolor=blue}
\begin{document}

\title{Free-hand Sketch Synthesis with Deformable Stroke Models
}
\subtitle{}


\author{Yi Li \and
        Yi-Zhe Song \and
        Timothy Hospedales \and
        Shaogang Gong
}


\institute{Yi Li \at
              \email{yi.li@qmul.ac.uk}           
           \and
           Yi-Zhe Song \at
              \email{yizhe.song@qmul.ac.uk}
           \and
           Timothy Hospedales \at
              \email{t.hospedales@qmul.ac.uk}
                         \and
           Shaogang Gong \at
              \email{s.gong@qmul.ac.uk}    
              \and
           School of Electronic Engineering and Computer Science, Queen Mary University of London, London, United Kingdoms
}

\date{Received: date / Accepted: date}

\maketitle

\begin{abstract}
We present a generative model which can automatically summarize the stroke composition of free-hand sketches of a given category. When our model is fit to a collection of 
sketches with similar poses, it discovers and learns the
structure and appearance of a set of coherent parts, with each part represented by a group of strokes. It represents
both consistent (topology) as well as diverse aspects (structure and appearance variations)
of each sketch category. Key to the success of our model are
important insights learned from a comprehensive study performed on
human stroke data. 
By fitting this model to images, we are able to
synthesize visually similar and pleasant free-hand sketches. 

\end{abstract}

\section{Introduction}
\begin{figure*}
\centering
\includegraphics[height=2.68in]{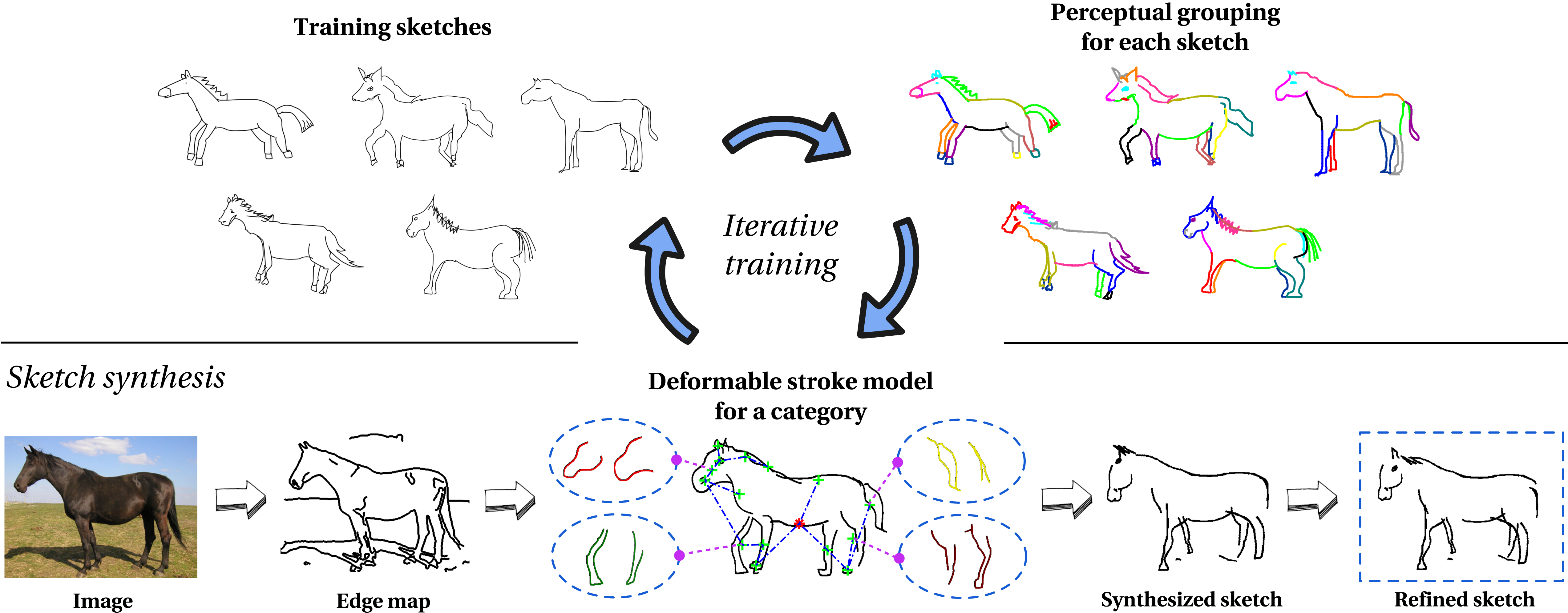}
   \caption{\label{fig:teaser}An overview of our framework, encompassing deformable stroke model (DSM) learning and free-hand sketch synthesis for given images. To learn a DSM, i) raw sketch strokes are grouped into semantic parts by perceptual grouping (semantic parts are not totally consistent across sketches); ii) category-level DSM is learned upon those semantic parts (category-level semantic parts are summarized and encoded); iii) the learned DSM is used to guide the perceptual grouping in the next iteration until convergence. When the DSM is obtained, we can synthesize sketches for given images, and the synthesized sketches from this model are highly similar to the original images and of a clear free-hand style. }  
\end{figure*}

Sketching comes naturally to humans. With the proliferation of touchscreens, we can now sketch effortlessly and ubiquitously by sweeping fingers on phones, tablets and smart watches. Studying free-hand sketches has thus become increasingly popular in recent years, with a wide spectrum of work addressing sketch recognition, sketch-based image retrieval, and sketching style and abstraction.  

While computers are approaching human level on recognizing free-hand
sketches (\cite{eitz2012hdhso,Schneider:2014:SCC:2661229.2661231,qianyu2015bmvc}), their capability of synthesizing sketches, especially free-hand sketches, has not been fully explored. The main existing works on sketch synthesis are engineered specifically and exclusively for a single category: human faces. Albeit successful at synthesizing sketches, important assumptions are ubiquitously made that render them not directly applicable to wider categories. It is often assumed that because faces exhibit quite stable structure (i) hand-crafted models specific to faces are sufficient to capture structural and appearance variations, (ii) auxiliary datasets of part-aligned photo and sketch pairs are mandatory and must be collected and annotated (however labour intensive), (iii) as a result of the strict data alignment, sketch synthesis is often performed in a relatively ad-hoc fashion, e.g., simple patch replacement. With a single exception that utilized professional strokes (rather than patches) (\cite{Berger:2013:SAP:2461912.2461964}), synthesized results resemble little the style and abstraction of free-hand sketches.

In this paper, going beyond just one object category, we present a generative data-driven model for free-hand sketch synthesis of diverse object categories. In contrast with prior art, (i) our model is capable of capturing structural and appearance variations without the handcrafted structural prior, (ii) we do not require purpose-built datasets to learn from, but instead utilize publicly available datasets of free-hand sketches that exhibit no alignment nor part labeling and (iii) our model optimally fits free-hand strokes to an image via a detection process, thus capturing the specific structural and appearance variation of the image and  performing synthesis in free-hand sketch style.

By training on a few sketches of similar poses (e.g., standing horse facing left), our model automatically discovers semantic parts -- including their number, appearance and topology -- from stroke data, as well as modeling their variability in appearance and location. For a given sketch category, we construct a deformable stroke model (DSM), that models the category at a stroke-level meanwhile encodes different structural variations (deformable). Once a DSM is learned, we can perform image to free-hand sketch conversion by synthesizing a sketch with the best trade-off between an image edge map and a prior in the form of the learned sketch model. This unique capability is critically dependent on our DSM that represents enough stroke diversity to match any image edge map, while simultaneously modeling topological layout so as to ensure visual plausibility.

Building such a model automatically is challenging. Similar models designed for images either require intensive supervision (\cite{Felzenszwalb:2005:PSO:1024426.1024429}) or produce imprecise and duplicated parts (\cite{Shotton:2008:MCO:1383053.1383268,Opelt06}). Thanks to a comprehensive analysis into stroke data that is unique to free-hand sketches, we demonstrate how semantic parts of sketches can be accurately extracted under minimum supervision. More specifically, we propose a perceptual grouping algorithm that forms raw strokes into semantically meaningful parts, which for the first time synergistically accounts for cues specific to free-hand sketches such as stroke length and temporal drawing order. The perceptual grouper enforces part semantics within an individual sketch, yet to build a category-level sketch model, a mechanism is required to extract category-level parts. For that, we further propose an iterative framework that interchangeably performs: (i) perceptual grouping on individual sketches, (ii) category-level DSM learning, and (iii) DSM matching/stroke labeling on training sketches. Once learned, our model generally captures all semantic parts shared across one object category without duplication. An overview of our work is shown in Figure \ref{fig:teaser}, including both deformable stroke model learning and the free-hand sketch synthesis application.

The contribution of our work is threefold : 
\begin{itemize}
\item A comprehensive and empirical analysis of sketch stroke data, highlighting the relationship between stroke length and stroke semantics, as well as the reliability of the stroke temporal order.

\item A perceptual grouping algorithm based on stroke analysis is proposed, which for the first time synergistically accounts for multiple cues, notably stroke length and stroke temporal order. 

\item By employing our perceptual grouping method, a deformable stroke model is automatically learned in an iterative process. This model encodes both the common topology and the variations in structure and appearance of a given sketch category. Afterwards a novel and general sketch synthesis application is derived from the learned sketch model.
\end{itemize}

We evaluate our framework via user studies and  experiments on two publicly available sketch datasets: (i) six diverse categories from non-expert sketches from the TU-Berlin dataset (\cite{eitz2012hdhso}) including: \emph{horse, shark, duck, bicycle, teapot} and \emph{face}, and (ii) professional sketches of two \emph{abstraction levels (90s and 30s)} of two artists in the Disney portrait dataset (\cite{Berger:2013:SAP:2461912.2461964}).

\section{Related work}
\label{sec:relwork}



\noindent\textbf{Data-driven sketch synthesis}\quad 
Early sketch synthesis models focus on broadening the gamut of styles with little
consideration paid to abstraction, thus producing sketches that
look like input photos
(\cite{Winkenbach:1994:CPI:192161.192184,Gooch:2004:HFI:966131.966133}). Later attempts convert images to sketch-like edge maps, which despite
being more abstract still closely resemble natural image statistics
(\cite{Guo:2007:PSI:1235884.1235965,yi2013icip}). Data-driven approaches have been
introduced to generate more human-like sketches, exclusively for one object category: human
faces. \cite{Chen:2002:PPI:641007.641040,Liang:2002:ECG:826030.826637}
took  simple exemplar-based approachs to synthesize faces
and used holistic training sketches. \cite{wang2009face,conf/cvpr/WangZLP12} decompose training
image-sketch pairs into patches, and train a  patch-level mapping model. All face synthesis systems above work with professional
sketches and assume perfect alignment across all training and testing data. As a
result, image and patch-level replacement strategies are often
sufficient to synthesize sketches. 

Moving onto free-hand sketches, \cite{Berger:2013:SAP:2461912.2461964} directly use strokes of a portrait sketch dataset collected from professional artists, and learn a set of parameters that reflect style and abstraction of different artists. They achieved this by building artist-specific stroke libraries and performing  stroke-level study on them with multiple characteristics accounted. Upon synthesis, they first convert image edges into vector curves according to a chosen style, then replace them with human strokes measuring shape, curvature and length. Although these stroke-level operations provided more freedom during synthesis, the assumption of rigorous alignment, in the form of manually fitting a face-specific mesh model to both images and sketches, is still made making extension to wider categories non-trivial. Their work laid a solid foundation for future study on free-hand sketch synthesis, yet extending it to many categories presents three major challenges: (i) sketches with fully annotated parts/feature points are difficult and costly to acquire, especially for more than one category; (ii) intra-category appearance and structure variations are larger in categories other than faces, and (iii) a better means of model fitting is required to account for noisier edges. In this paper, we design a model that is flexible enough to account for all these highlighted problems.

\noindent\textbf{Contour models and pictorial structure}\quad  Our model is inspired by contour (\cite{Shotton:2008:MCO:1383053.1383268,Opelt06,Ferrari2010,Dai2013})
and pictorial structure models (\cite{Felzenszwalb:2005:PSO:1024426.1024429}). Both have been shown to work well in the image domain, especially in terms of addressing holistic structural variation and noise robustness. The idea behind contour models is learning object parts directly on edge fragments. And a by-product of the contour model is that via detection an instance of the model will be left on the input image. Despite being able to generate sketch-like instances of the model, the main focus of that work is on object detection, therefore synthesized results do not exhibit sufficient aesthetic quality. Major drawbacks of contour models in the context of sketch synthesis are: (i) duplicated parts and missing details as a result of unsupervised learning, (ii) rigid star-graph structure and relatively weak detector are not good at modeling sophisticated topology and enforcing plausible sketch geometry, and (iii) inability to address appearance variations associated with local contour fragments. On the other hand, pictorial structure models are very efficient at explicitly and accurately modeling all mandatory parts and their spatial relationships. They work by using a minimum spanning tree  and casting model learning and detection into a statistical maximum a posteriori (MAP) framework. Yet the much desired model accuracy is achieved at the cost of supervised learning that involves intensive labeling a priori.

By integrating pictorial structure and contour models, we propose a deformable stroke model that: (i) employs perceptual grouping and an iterative learning scheme, and thus yields accurate models with minimum human effort, (ii)  customizes the model learning and detection framework of pictorial structure to address more sophisticated topology possessed by sketches and achieve more effective stroke to edge map registration, and (iii) augments contour model parts from just one uniform contour fragment to multiple stroke exemplars in order to capture local appearance variations.

\noindent\textbf{Stroke analysis}\quad Despite the recent surge in sketch research, stroke-level analysis of human sketches remains sparse. Existing studies (\cite{eitz2012hdhso,Berger:2013:SAP:2461912.2461964,Schneider:2014:SCC:2661229.2661231}) have mentioned stroke ordering, categorizing strokes into types, and the importance of individual strokes for recognition. However, a detailed analysis has been lacking especially towards: (i) level of semantics encoded by human strokes, and (ii) the temporal sequencing of strokes within a given category. 

 \cite{eitz2012hdhso} proposed a dataset of 20,000 human sketches and offered anecdotal evidence towards the role of stroke ordering.  \cite{Fu:2011:ACL:2070781.2024167} claimed that human generally sketch in a hierarchical fashion, i.e., contours first, details second. Yet as can be seen later in Section~\ref{sec:stroke}, we found this  does not always hold, especially for non-expert sketches. More recently, \cite{Schneider:2014:SCC:2661229.2661231} touched on stroke importance and demonstrated empirically that certain strokes are more important for sketch recognition. While interesting, none of the work above provided means of modeling stroke ordering/saliency inside a computational framework, thus making potential applications unclear. \cite{SketchSegmentationLabeling:2014} was first in actually utilizing temporal ordering of strokes as a soft grouping constraint. Similar to them, we also employ stroke ordering as a cost term in our grouping framework. Yet while they only took the temporal order grouping cue as a hypothesis, we move on to provide solid evidence to support this usage.

A more comprehensive analysis of strokes was performed by \cite{Berger:2013:SAP:2461912.2461964} aiming to decode the style and abstraction of different artists. 
They claimed that stroke length correlates positively with abstraction level, and in turn categorized strokes into several types based on their geometrical characteristics. Although insightful, their analysis was constrained to a dataset of professional portrait sketches, whereas we perform an in-depth study into non-expert sketches of many categories as well as the professional portrait dataset and we specifically aim to understand stroke semantics rather than style and abstraction. 

\noindent\textbf{Perceptual grouping of strokes}\quad
\cite{SketchSegmentationLabeling:2014} remains the single study on stroke grouping to date. They worked with sketches of 3D objects, assuming that sketches do not possess noise or over-sketching (obvious overlapping strokes). Instead, we work on free-hand sketches where noise and over-sketching are pervasive. Informed by a stroke-level analysis, our grouper not only uniquely considers temporal order and several Gestalt principles, but also controls group size to ensure semantic meaningfulness. Beside applying it on individual sketches, we  also integrate the grouper with stroke model learning to achieve across-category consistency. 

\section{Stroke analysis}\label{sec:stroke}

In this section we perform a full analysis on how stroke-level information can be best used to locate semantic parts of sketches. In particular, we look into (i) the correlation between stroke length and its semantics as an object part, i.e., what kind of strokes do object parts correspond to, and (ii) the reliability of temporal ordering of strokes as a grouping cue, i.e., to what degree can we rely on temporal information of strokes. We conduct our study on both non-expert and professional sketches: (i) six diverse categories from non-expert sketches from the TU-Berlin dataset (\cite{eitz2012hdhso}) including: \emph{horse, shark, duck, bicycle, teapot} and \emph{face}, and (ii) professional sketches of two \emph{abstraction levels (90s and 30s)} of \emph{artist A} and \emph{artist E} in the Disney portrait dataset (\cite{Berger:2013:SAP:2461912.2461964}).

\begin{figure}[tb]
\centering
\includegraphics[scale=0.165]{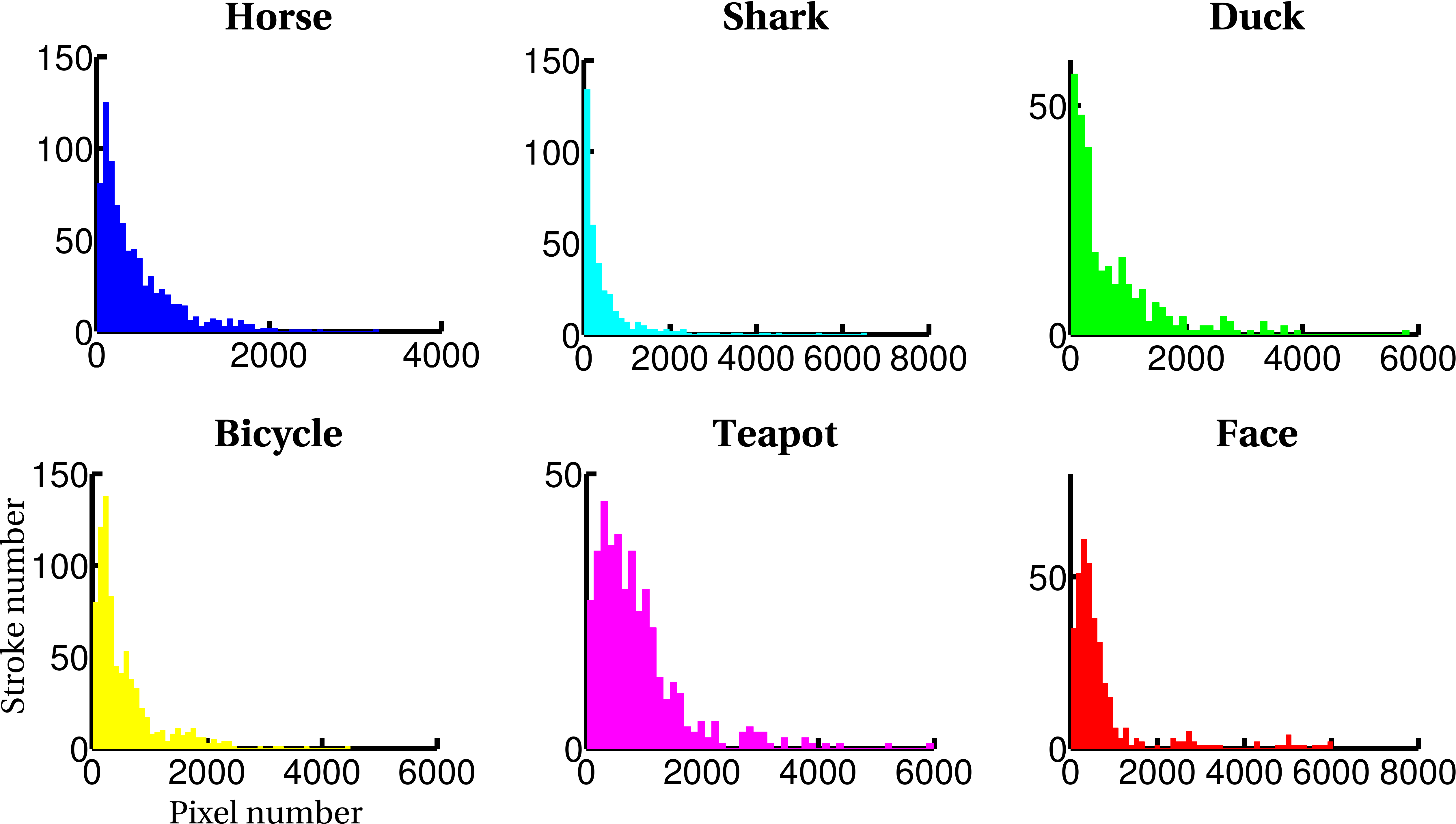}
\caption{\label{fig:strlen1}Histograms of stroke lengths of six non-expert sketch categories. (x-axis: the size of stroke in pixels; y-axis:  number of strokes in the category). }
\end{figure}

\begin{figure}[tb]
\centering
\subfigure[\label{fig:strlen2}]{
\includegraphics[scale=0.11]
{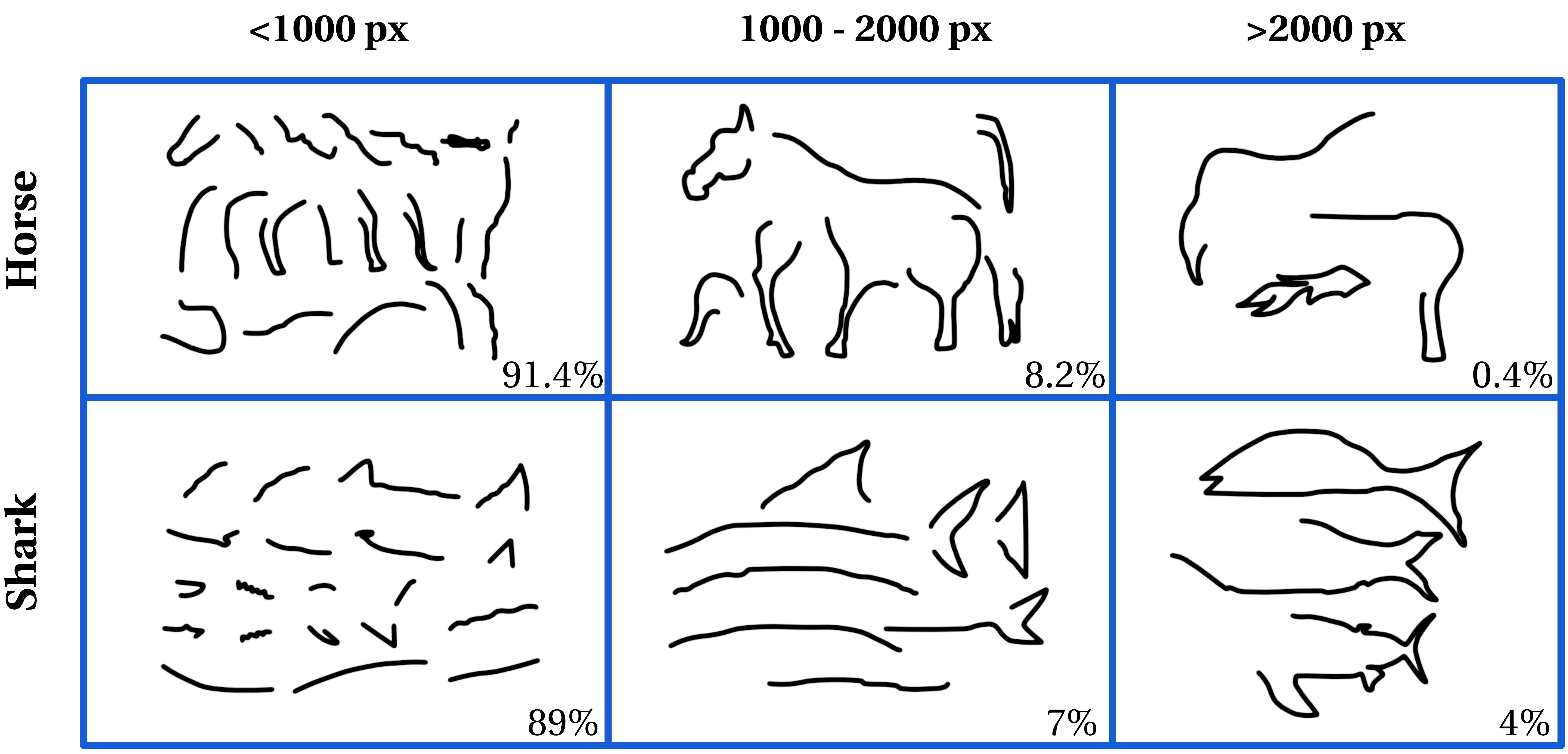}
}
\subfigure[\label{fig:strlen3}]{
\includegraphics[scale=0.11]
{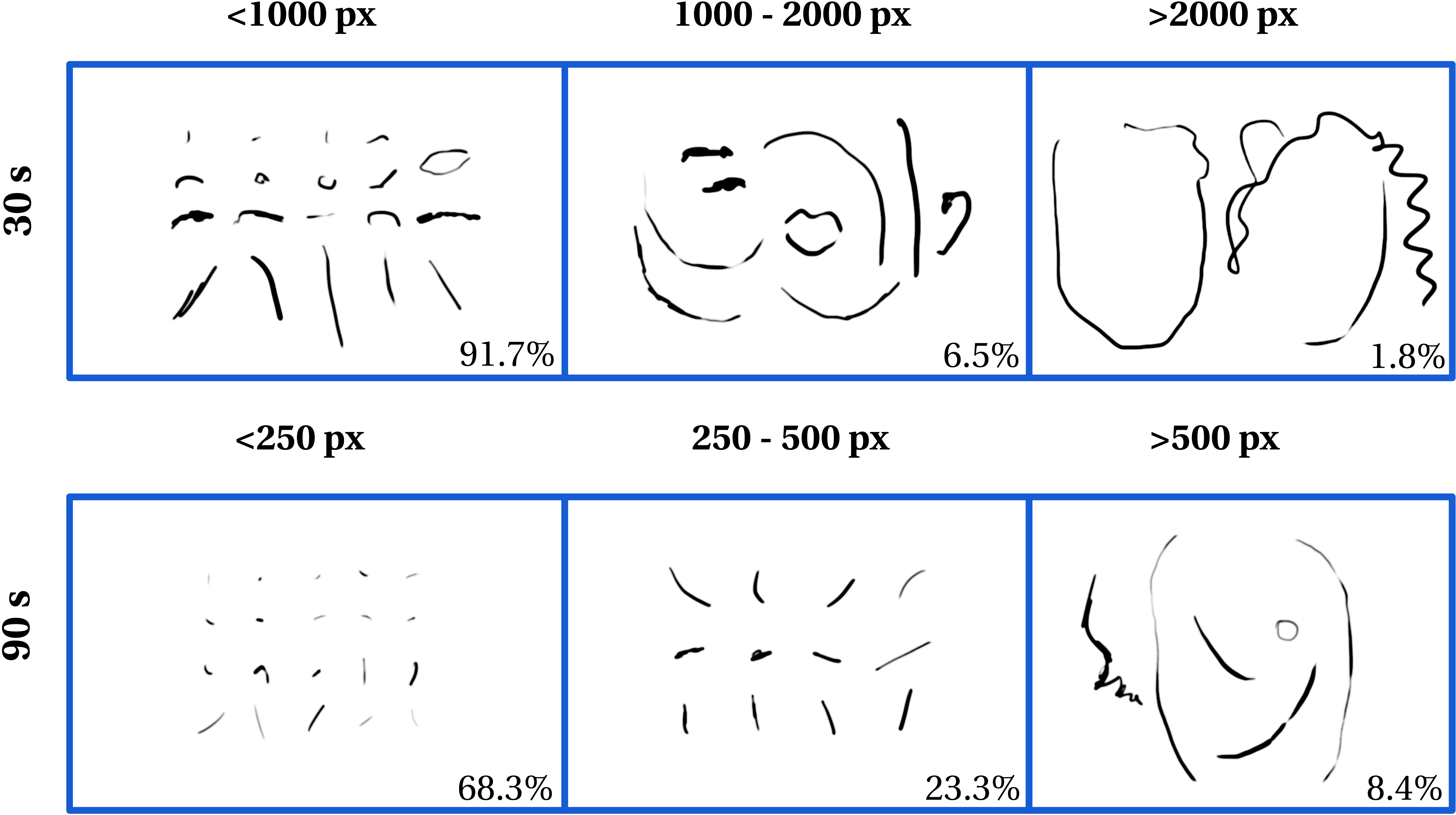}
}
\caption{\label{fig:strlen} Example strokes of each size group. (a) 2 categories in TU-Berlin dataset. (b) 2 levels of abstraction from artist A in Disney portrait dataset. The proportion of each size group in the given category is indicated in the bottom-right corner of each cell.}
\end{figure}

\noindent\textbf{Semantics of strokes}\quad 
On the TU-Berlin dataset, we first measure stroke length statistics (quantified by pixel count) of all six chosen categories. Histograms of each category are provided in Figure \ref{fig:strlen1}. It can be observed that despite minor cross-category variations, distributions are always long-tailed: most strokes being shorter than 1000 pixels, with a small proportion exceeding 2000 pixels. We further divide strokes into 3 groups based on length,  illustrated by examples of 2 categories in Figure~\ref{fig:strlen2}. We can see that (i) medium-sized strokes tend to exhibit semantic parts of objects, (ii) the majority of short strokes (e.g., $<1000$ px) are too small to correspond to a clear part, and (iii) long strokes (e.g., $>2000$ px) lose clear meaning by encompassing more than one semantic part. 

These observations indicate that, ideally, a stroke model can be directly learned on strokes from the medium length range. However, in practice, we further observe that people tend to draw very few medium-sized strokes (length correlates negatively with quantity as seen in Figure \ref{fig:strlen1}), making them statistically insignificant for model learning. This is apparent when we look at percentages of strokes in each range, shown towards bottom right of each cell in Figure \ref{fig:strlen1}. We are therefore motivated to propose a perceptual grouping mechanism that counters this problem by grouping short strokes into longer chains that constitute object parts (e.g., towards the medium range in the TU-Berlin sketch dataset). We call the grouped strokes representing semantic parts as semantic strokes. Meanwhile, a cutting mechanism is also employed to process the few very long strokes into segments of short and/or medium length, which can be processed by perceptual grouping afterwards. 

On the Disney portrait dataset, a statistical analysis of strokes similar to Figure \ref{fig:strlen1} was already  conducted by the original authors and the stroke length distributions are quite similar to ours. From example strokes in each range in Figure \ref{fig:strlen3}, we can see for sketches of the $30$s level the situation is similar to the TU-Berlin dataset where most semantic strokes are clustered within the middle length range (i.e., $1000-2000$ px) and the largest group is still the short strokes. As already claimed in \cite{Berger:2013:SAP:2461912.2461964} and also reflected in the bottom row of Figure \ref{fig:strlen3}, stroke lengths across the board reduce significantly as abstraction level goes down to $90$s. This suggests that, for the purpose of extracting semantic parts, a grouping framework is even more necessary for professional sketches where individual strokes convey less semantic meaning.

\begin{figure*}[htb]
\centering
\includegraphics[scale=0.255]{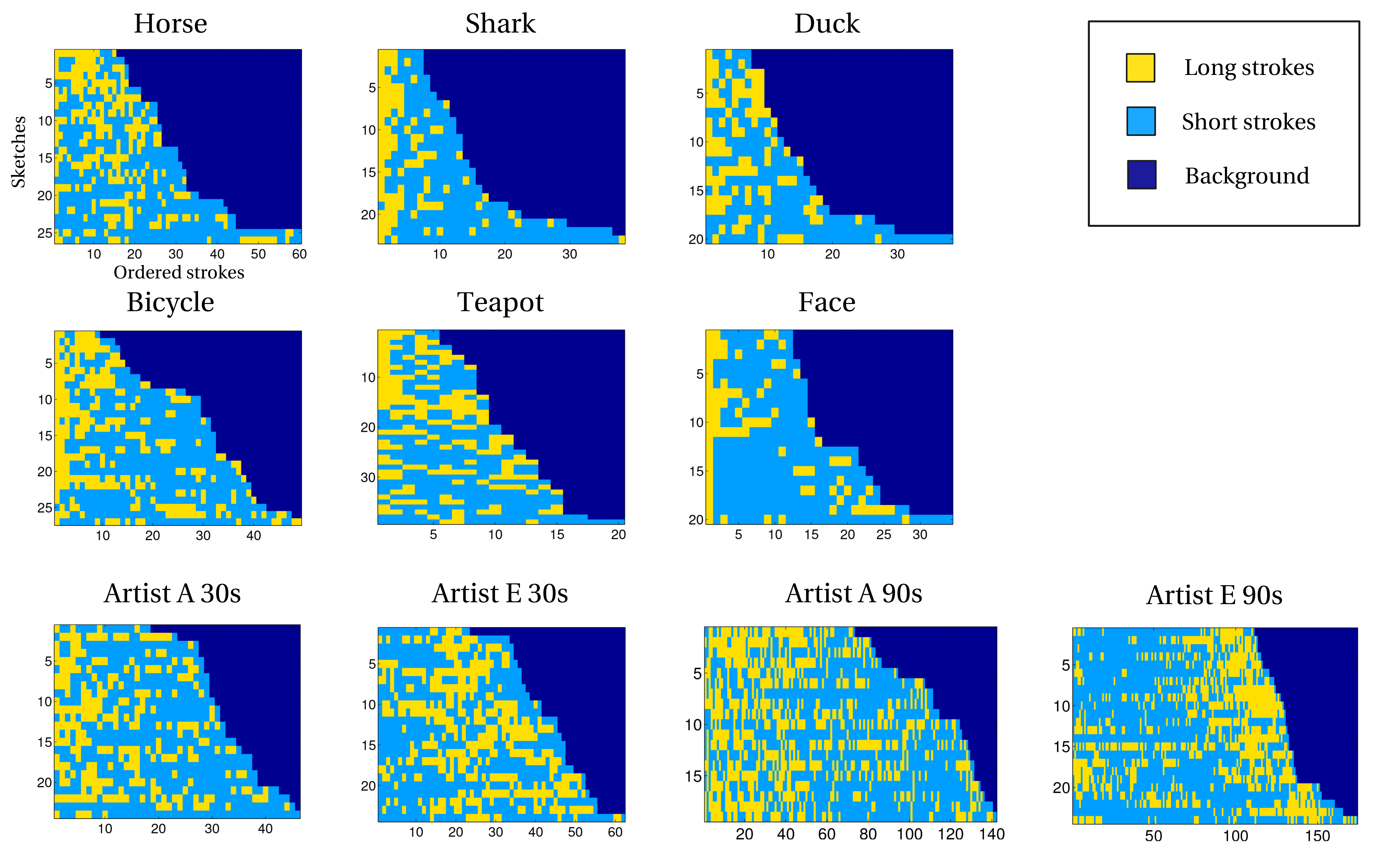}
\caption{\label{fig:temp}Exploration of stroke temporal order. Subplots represent 10 categories: \emph{horse}, \emph{shark}, \emph{duck}, \emph{bicycle}, \emph{teapot} and \emph{face} of TU-Berlin dataset and $30$s and $90$s levels of \emph{artist A} and \emph{artist E} in Disney portrait dataset. x-axis shows stroke order and y-axis sketch samples, so each cell of the matrices is a stroke. Sketch samples are sorted by their number of strokes (abstraction). Shorter than average strokes are yellow, longer than average strokes are cyan.}
\end{figure*}

\noindent\textbf{Stroke ordering}\quad 
Another previously under-studied cue for sketch understanding is the temporal ordering of strokes, with only a few studies exploring this (\cite{Fu:2011:ACL:2070781.2024167,SketchSegmentationLabeling:2014}). Yet these authors only hypothesized the benefits of temporal ordering without critical analysis a priori. In order to examine if there is a consistent trend in holistic stroke ordering (e.g., if long strokes are drawn first followed by short strokes), we color-code length of each stroke in Figure~\ref{fig:temp} where: each sketch  is represented by a row of colored cells, ordering along the x-axis reflects drawing order, and sketches (rows) are sorted in ascending order of number of constituent strokes. For ease of interpretation, only 2 colors are used for the color-coding. Strokes with above average length are encoded as yellow and those with below average as cyan.

\begin{figure}[tb]
\centering
\includegraphics[scale=0.12]{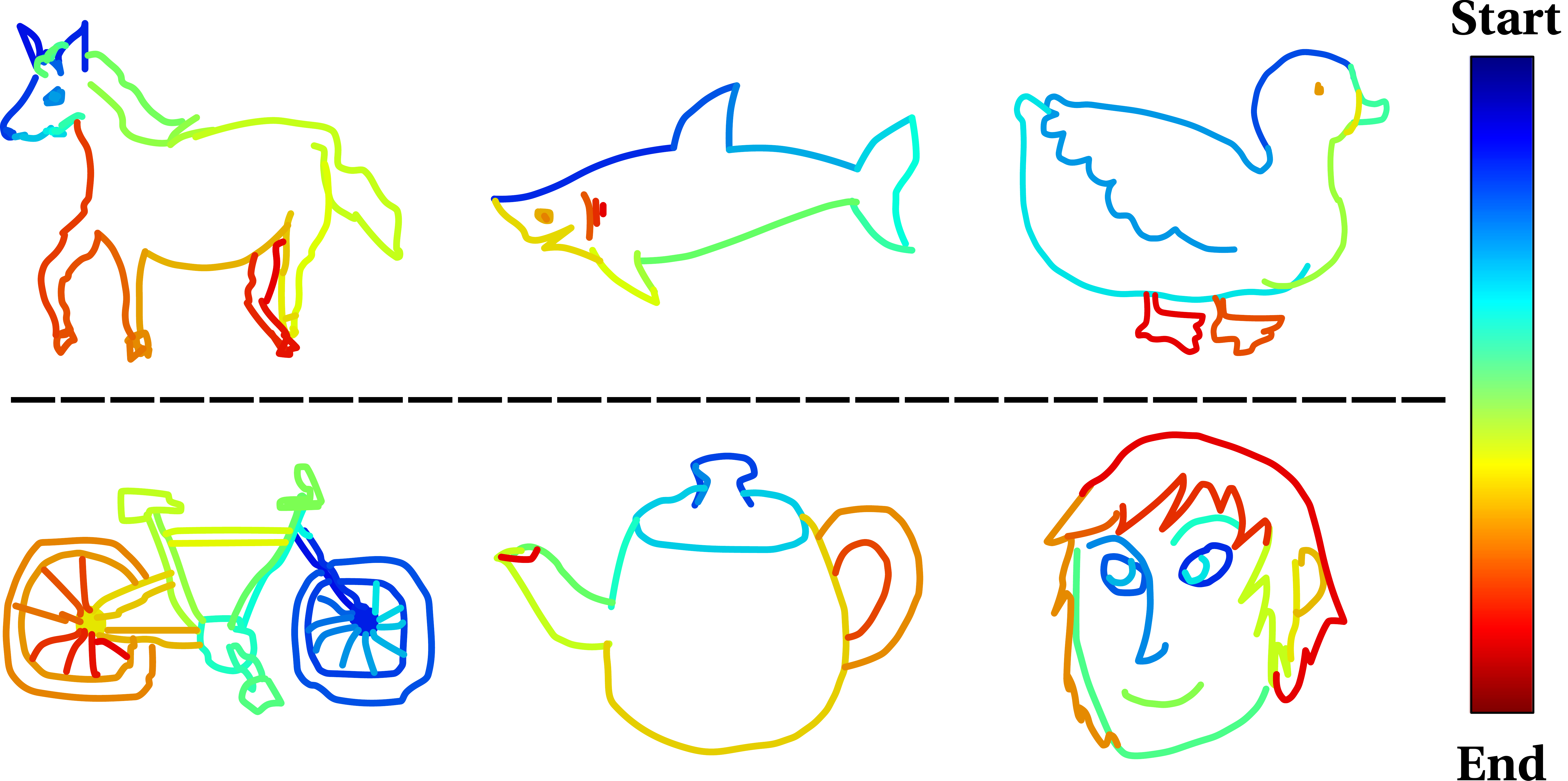}
\caption{\label{fig:orders}Stroke drawing order encoded by color (starts from blue and ends at red). Object parts tend to be drawn with sequential strokes.}
\end{figure}

From Figure~\ref{fig:temp} (1st and 2nd rows), we can see that non-expert sketches with fewer strokes tend to contain a bigger proportion of longer strokes (greater yellow proportion in the upper rows), which matches the claim made by \cite{Berger:2013:SAP:2461912.2461964}. However, there is not a clear trend in the ordering of long and short strokes across all the categories. Although clearer trend of short strokes following long strokes can be observed in few categories, e.g., \emph{shark} and \emph{face}, and this is due to these categories' contour can be depicted by very few long and simple strokes. In most cases, long and short strokes appear interchangeably at random. Only in the more abstract sketches (upper rows), we can see a slight trend of long strokes being used more towards the beginning (more yellow on the left). This indicates that average humans draw sketches with a random order of strokes of various lengths, instead of a coherent global order in the form of a hierarchy (such as long strokes first, short ones second). In Figure~\ref{fig:temp} (3rd row), we can see that artistic sketches exhibit a clearer pattern of a long stroke followed by several short strokes (the barcode pattern in the figure). However, there is still not a dominant trend that long strokes in general are finished before short strokes. This is different from the claim made by \cite{Fu:2011:ACL:2070781.2024167}, that most drawers, both amateurs and professionals, depict objects hierarchically. In fact, it can also be observed from Figure~\ref{fig:orders} that average people often sketch objects part by part other than hierarchically. However the ordering of how parts are drawn appears to be random.
 
Although stroke ordering shows no global trend, we found that local stroke ordering (i.e., strokes depicted within a short timeframe) does possess a level of consistency that could be useful for semantic stroke grouping. Specifically, we observe that people tend to draw a series of consecutive strokes to depict one semantic part, as seen in Figure~\ref{fig:orders}. The same hypothesis was also made by \cite{SketchSegmentationLabeling:2014}, but without clear stroke-level analysis beforehand. Later, we will demonstrate via our grouper how local temporal ordering of strokes can be modeled and help to form semantic strokes. 

\section{A deformable stroke model}
From a collection of sketches of similar poses within one category, we can learn a generative deformable stroke model (DSM). 
In this section, we first formally define DSM. Then, we introduce the perceptual grouping which groups raw strokes into semantic strokes/parts, and we illustrate how a DSM is learned on those semantic parts and how to use DSM to detect on sketches/images. Finally, the iterative process of performing these three steps interchangeably is well demonstrated with concrete examples.

\begin{figure*}[tb]
\centering
\includegraphics[width=2.\columnwidth]{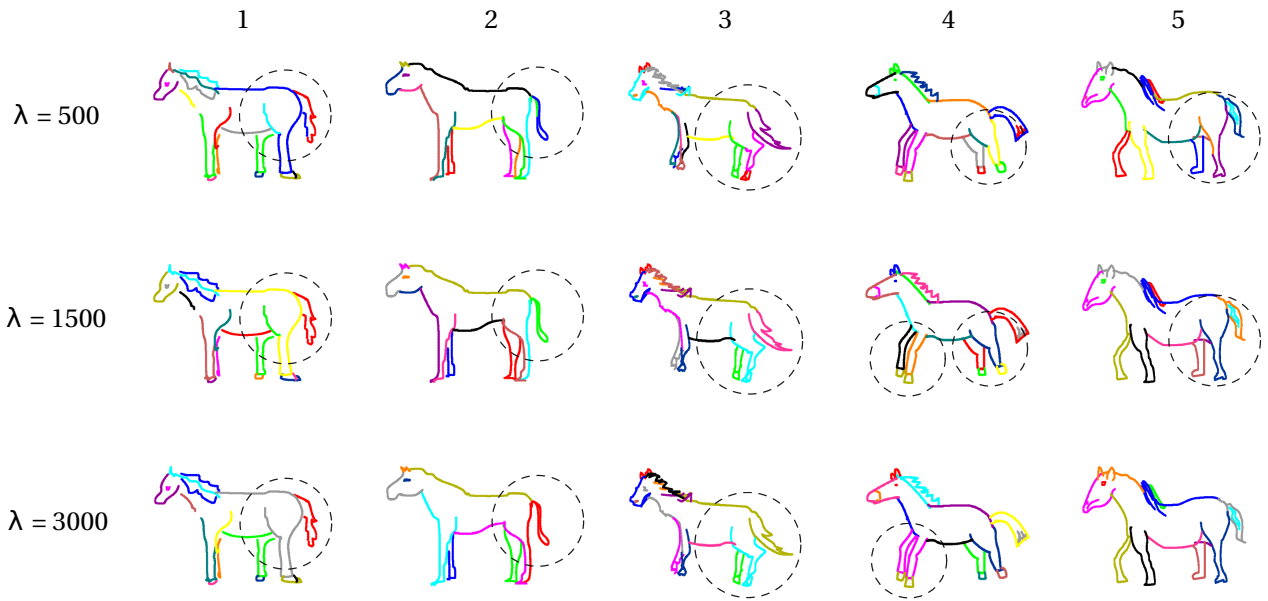}
\caption{\label{fig:len} The effect of changing $\lambda$ to control the semantic stroke length (measured in pixels). We can see as $\lambda$ increases, the semantic strokes' lengths increase as well. And generally speaking, when a proper semantic length is set, the groupings of the strokes are more semantically proper (neither over-segmented or over-grouped). More specifically, we can see that when $\lambda=500$, many tails and back legs are fragmented. But when $\lambda=1500$, those tails and back legs are grouped much better. Beyond that, when $\lambda=3000$, two more semantic parts tend to be grouped together improperly, e.g., one back leg and the tail (column 2), the tail and the back (column 3), or two front legs (column 4). Yet it can also be noticed that when a horse is relatively well drawn (each part is very distinguishable), the stroke length term will influence less, e.g., column 5.  }
\end{figure*}

\subsection{Model definition}
\label{sec:moddef}
Our DSM is an undirected graph of $n$ semantic part clusters: $G=(V,E)$. The vertices $V=\{v_1,...,v_n\}$ represent category-level semantic part clusters, and pairs of semantic part clusters are connected by an edge $(v_i,v_j)\in E$ if their locations are closely related. The model is parameterized by $\theta=(u,E,c)$, where $u=\{u_1,...,u_n\}$, with $u_i=\{s_i^a\}_{a=1}^{m_i}$ representing $m_i$ semantic stroke exemplars of the semantic part cluster $v_i$; $E$ encodes pairwise part connectivity; 
and $c=\{c_{ij}|(v_i,v_j)\in E\}$ encodes the spatial relation between connected part clusters. An example \emph{shark} DSM illustration with full part clusters is shown in Figure~\ref{fig:elem} (and a partial example for \emph{horse} is already shown in Figure~\ref{fig:teaser}), where the green crosses are the vertices $V$ and the blue dashed lines are the edges $E$. The part exemplars $u_i$ are highlighted in blue dashed ovals. 

\subsection{Perceptual grouping}\label{sec:percepGroup}
Perceptual grouping creates the building blocks (\emph{semantic strokes/parts}) for model learning based on \emph{raw stroke} input. There are many factors that need to be considered in perceptual grouping.
As demonstrated in Section~\ref{sec:stroke}, small strokes need to be grouped to be semantically meaningful, and local temporal order is helpful to decide whether strokes are semantically related. Equally important to the above, conventional perceptual grouping principles (Gestalt principles, e.g. proximity, continuity, similarity) are also required to decide if a stroke set should be grouped. Furthermore, after the first iteration, the learned DSM model is able to assign a group label for each stroke, which can be used in the next grouping iteration.


Algorithmically, our perceptual grouping approach is inspired by \cite{Barla:2005:GCL:1187112.1187227}, who iteratively and greedily group pairs of lines with minimum error. However, their cost function includes only proximity and continuity; and their purpose is line simplification, so grouped lines are replaced by new combined lines. We adopt the idea of iterative grouping but change and expand their error metric to suit our task. For grouped strokes, each stroke is still treated independently, but the stroke length is updated with the group length. 

More specifically, for each pair of strokes $s_1,s_2$, grouping error is calculated based on 6 aspects: proximity, continuity, similarity, stroke length, local temporal order and model label (only used from second iteration), and the error metric function is defined as:
\begin{multline}
\label{equ:pegr}
M(s_i,s_j)= (\omega_{pro} * D_{pro}(s_i,s_j)+\omega_{con} * D_{con}(s_i,s_j)\\+\omega_{len} * D_{len}(s_i,s_j)- \omega_{sim} * B_{sim}(s_i,s_j))\\ *F_{temp}(s_i,s_j)* F_{mod}(s_i,s_j),
\end{multline}
where proximity $D_{pro}$, continuity $D_{con}$ and stroke length $D_{len}$ are treated as cost/distance which increase the error, while similarity $B_{sim}$ decreases the error. Local temporal order $F_{temp}$ and model label $F_{mod}$ further modulate the overall error. All the terms have corresponding weights $\{\omega\}$, which make the algorithm cutomizable for different datasets. Detailed definitions and explanations for the 6 terms are as follows (to be noticed, as our perceptual grouping method is an unsupervised and greedy algorithm, the colors for the perceptual grouping results are just for differentiating grouped semantic strokes in individual sketches and have no correspondence between sketches):


\noindent\textbf{Proximity}\quad Proximity employs the modified Hausdorff distance (MHD) (\cite{Dubuisson94amodified}) $d_H(\cdot)$ between two strokes, which represents the average closest distance between two sets of edge points. We define $D_{pro}(s_i,s_j)=d_H(s_i,s_j)/\epsilon_{pro}$, dividing the calculated MHD with a factor $\epsilon_{pro}$ to control the scale of the expected proximity. Given the image size $\phi$ and the average semantic stroke number $\eta_{avg}$ of the previous iteration (the average raw stroke number for the first iteration),
we use $\epsilon_{pro}=\sqrt{\phi/\eta_{avg}}/2$, which roughly indicates how closely two semantically correlated strokes should be located.


\noindent\textbf{Continuity}\quad To compute continuity, we first find the closest endpoints $x$,$y$ of the two strokes. For the endpoints $x$,$y$, another two points $x'$,$y'$ on the corresponding strokes with very close distance (e.g., 10 pixels) to $x$,$y$ are also extracted to compute the connection angle. Finally, the continuity is computed as: 
\begin{equation*}
D_{con}(s_i,s_j) = \lVert x-y \rVert * ( 1 + angle(\overrightarrow{x'x},\overrightarrow{y'y}))/\epsilon_{con},
\end{equation*}
where $\epsilon_{con}$ is used for scaling, and set to $\epsilon_{pro}/4$, as continuity should have more strict requirement than the proximity.

\begin{figure}[tb]
\centering
\includegraphics[width=0.9\columnwidth]{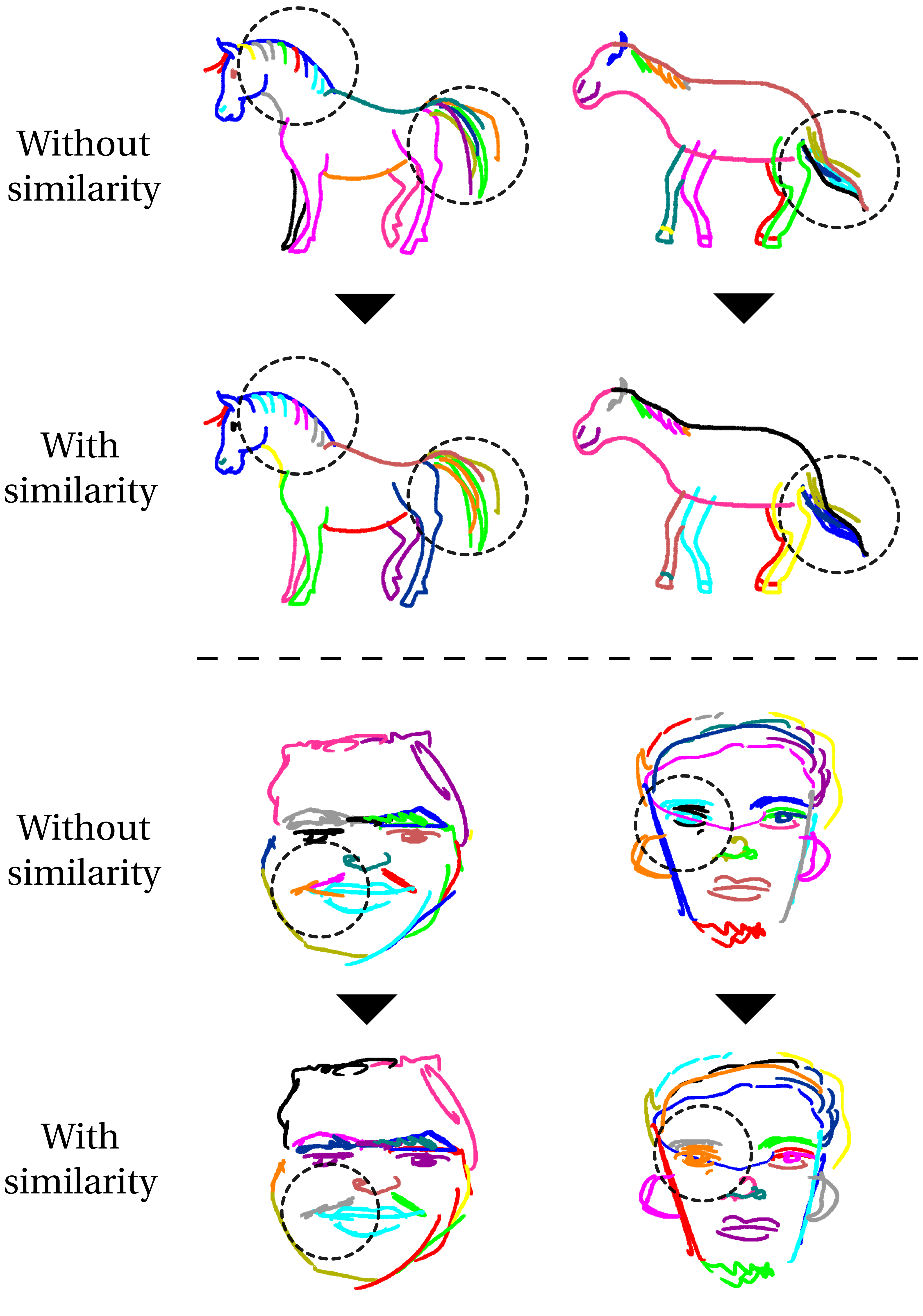}
\caption{\label{fig:simi} The effect of employing the similarity term. Many separate strokes or wrongly grouped strokes are correctly grouped into properer semantic strokes when exploiting similarity.}
\end{figure}

\noindent\textbf{Stroke length}\quad Stroke length cost is the sum of the length of the two strokes: $D_{len}(s_i,s_j)=(P(s_i)+P(s_j))/\lambda$,
where $P(s_i)$ is the length (pixel number) of raw stroke $s_i$; or if $s_i$ is already within a grouped semantic stroke, it is the stroke group length. 
The normalization factor is computed as $\lambda=\tau*\eta_{sem}$, where $\eta_{sem}$ is the estimated average number of strokes composing a semantic group in a dataset (from the analysis). When $\eta_{sem}=1$, $\tau$ is the proper length for a stroke to be semantically meaningful (e.g. around 1500 px in Figure~\ref{fig:strlen2}), and when $\eta_{sem}>1$, $\tau$ is the maximum length of all the strokes.

The effect of changing $\lambda$ to control the semantic stroke length is demonstrated in Figure~\ref{fig:len}.

\noindent\textbf{Similarity}\quad In some sketches, repetitive short strokes are used to draw texture like hair or mustache. Those strokes convey a complete semantic stroke, yet can be clustered into different groups by continuity. To correct this, we introduce a similarity bonus. We extract  strokes $s_1$ and $s_2$'s  shape context descriptor and calculate their matching cost $K(s_i,s_j)$ according to \cite{Belongie:2002:SMO:628328.628792}. The similarity bonus is then $B_{sim}(s_i,s_j)=exp(-K(s_i,s_j)^2/\sigma^2)$
where $\sigma$ is a scale factor. 
Examples in Figure~\ref{fig:simi} demonstrate the effect of this term.

\begin{figure}[tb]
\centering
\includegraphics[scale=0.078]{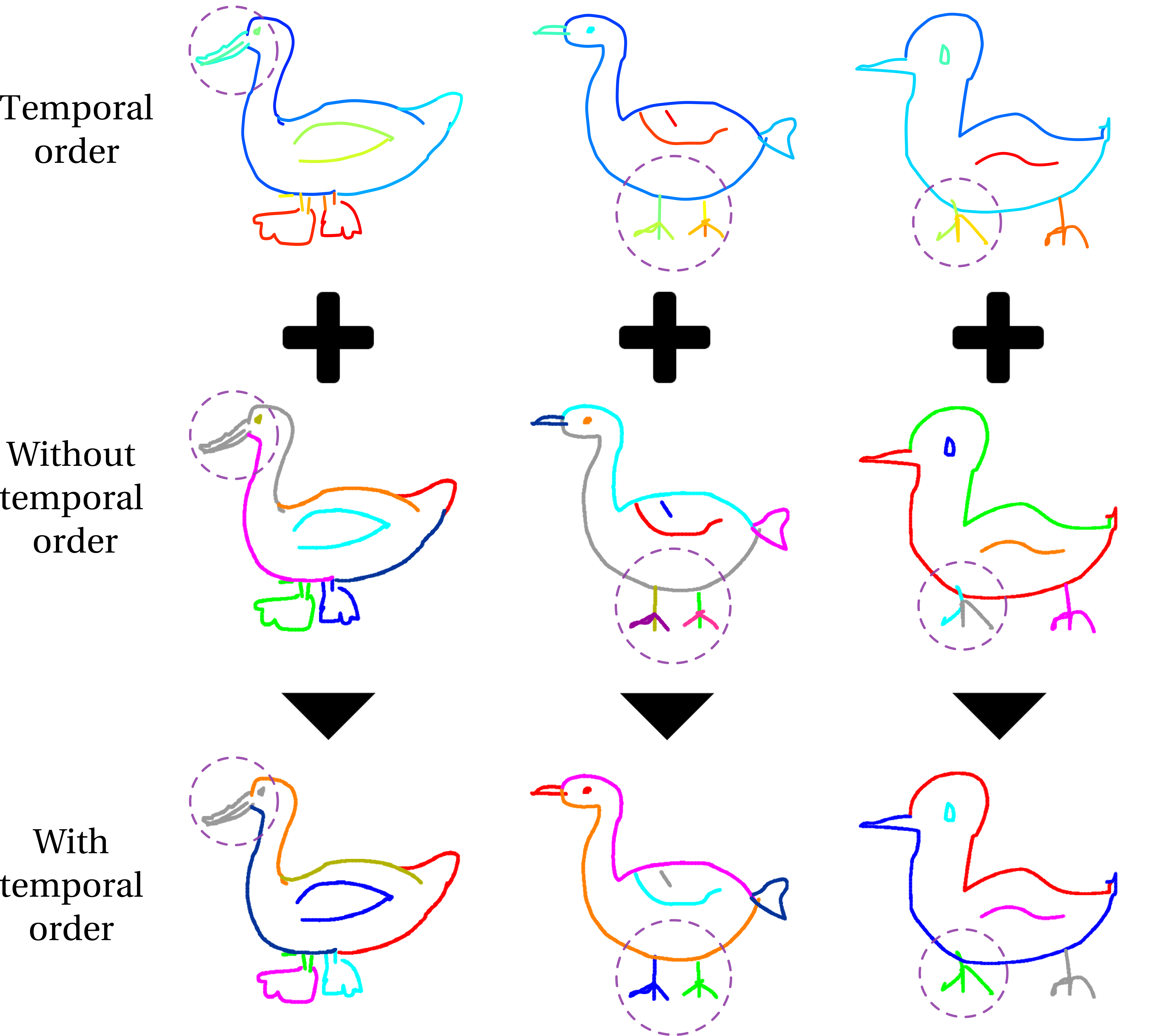}
\caption{\label{fig:temporal}The effect of employing stroke temporal order. We can see many errors made to the beak and feet (wrongly grouped with other semantic part or separated into several parts) are corrected as a result.}
\end{figure}

\noindent\textbf{Local temporal order}\quad The local temporal order provides an adjustment factor $F_{temp}$ to the previously computed error $M(s_i,s_j)$ based on how close the drawing orders of the two strokes are:
\begin{equation*}
F_{temp}(s_i,s_j)=\begin{cases}
    1-\mu_{temp}, & \text{if $|T(s_i) - T(s_j)|<\delta$}.\\
    1+\mu_{temp}, & \text{otherwise}.
  \end{cases},
\end{equation*}
where $T(s)$ is the order number of stroke $s$. $\delta=\eta_{all}/\eta_{avg}$ is the estimated maximum order difference in stroke order within a semantic stroke, where $\eta_{all}$ is the overall stroke number in the current sketch. $\mu_{temp}$ is the adjustment factor. The effect by this term is demonstrated in Figure~\ref{fig:temporal}.

\begin{figure}[tb]
\centering
\includegraphics[scale=0.092]{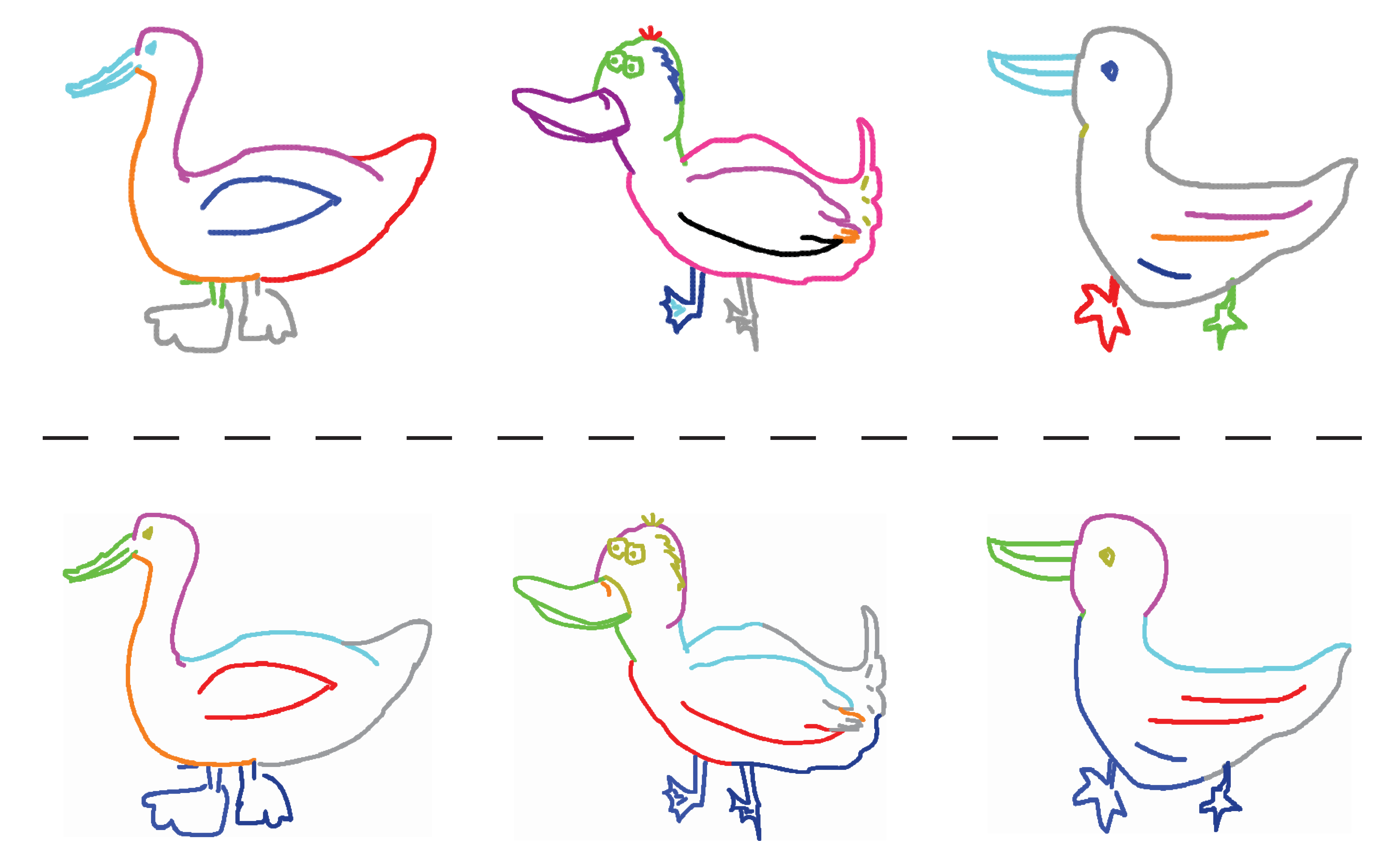}
\caption{\label{fig:model}The model label after the perceptual grouping of the first iteration. Above: first iteration perceptual groupings. Below: model labels. It can be observed that the first iteration perceptual groupings have different number of semantic strokes, and the divisions over the eyes, head and body are quite different across sketches. However, after a category-level DSM is learned, the model labels the sketches in a very similar fashion, roughly dividing the duck into beak(green), head(purple), eyes(gold), back(cyan), tail(grey), wing(red), belly(orange), left foot(light blue), right foot(dark blue). But some errors still exist in the model label, e.g., missing parts and wrongly labeled part, which will be further corrected in the future iterations.}
\end{figure}

\noindent\textbf{Model label}\quad The DSM model label provides a second adjustment factor according to whether two strokes have the same label or not. 
\begin{equation}
F_{mod}(s_i,s_j)=\begin{cases}
    1-\mu_{mod}, & \text{if $W(s_i) == W(s_j)$}.\\
    1+\mu_{mod}, & \text{otherwise}.
  \end{cases}\label{eq:modelLabel},
\end{equation}
where $W(s)$ is the model label for stroke $s$, and $\mu_{mod}$ is the adjustment factor.
The model label obtained after first iteration of perceptual grouping is shown in Figure~\ref{fig:model}.

\begin{figure}[tb]
\centering
\includegraphics[width=1.0\columnwidth]{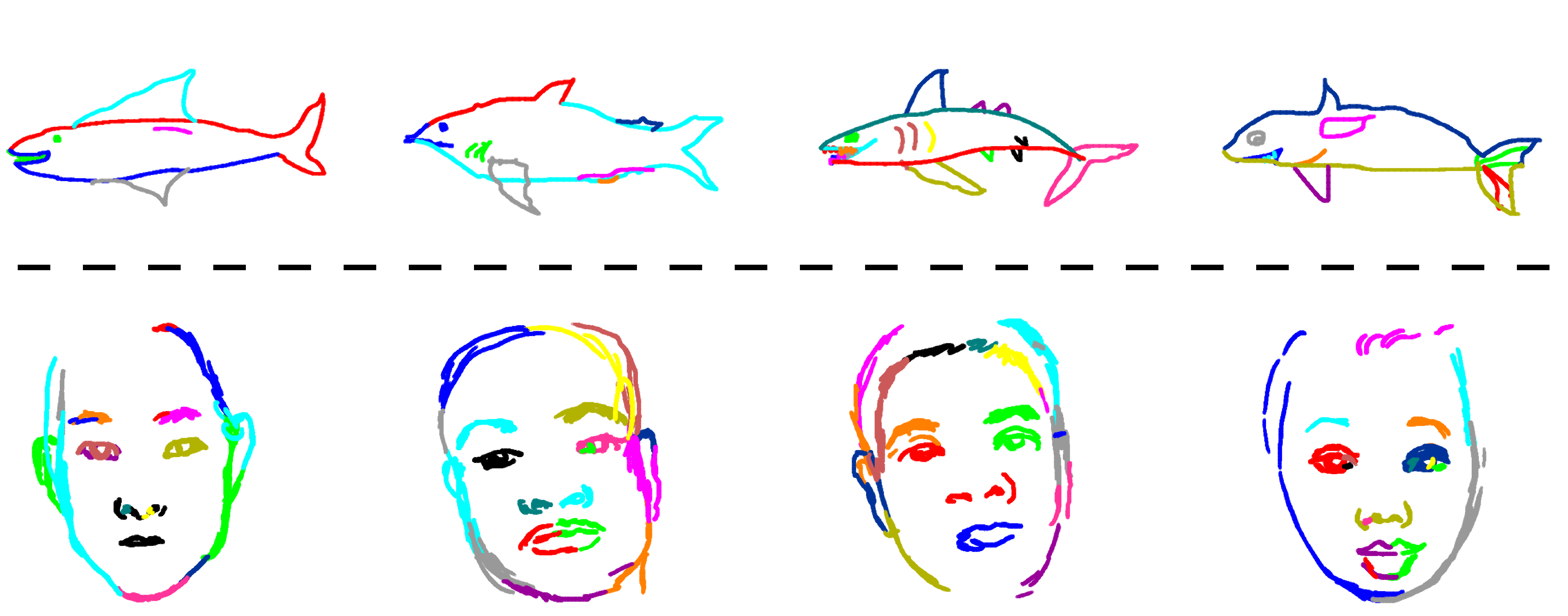}
\caption{\label{fig:grouping}Perceptual grouping results. For each sketch, a semantic stroke are represented by one color. }
\end{figure}

\begin{algorithm}[htb]
\caption{Perceptual grouping algorithm}
\label{alg:pegr}
\begin{algorithmic}
\State Input $t$ strokes $\{s_i\}_{i=1}^t$
\State Set the maximum error threshold to $h$
\For{$i,j=1 \to t$}
\State $ErrorMx(i,j) = M(s_i,s_j)$  \Comment{Pairwise error matrix}
\EndFor
\While 1
\State $[s_a,s_b,minError]=\min(ErrorMx)$\\
\Comment{Find $s_a,s_b$ with the smallest error}
\If{$minError==h$}
\State $break$
\EndIf
\State $ErrorMx(a,b) \gets h$
\If{None of $s_a,s_b$ is grouped yet}
\State Make a new group and group $s_a,s_b$
\ElsIf{One of $s_a,s_b$ is not grouped yet}
\State Group $s_a,s_b$ to the existing group
\Else
\State $continue$
\EndIf
\State Update $ErrorMx$ cells that are related to strokes in the current group according to the new group length
\EndWhile
\State Assign each orphan stroke a unique group id
\end{algorithmic}
\end{algorithm}
Pseudo code for our perceptual grouping algorithm is shown in Algorithm~\ref{alg:pegr}. More results produced by first iteration perceptual grouping are illustrated in Figure~\ref{fig:grouping}. As can be seen, every sketch is grouped into a similar number of parts, and there is reasonable group correspondence among the sketches in terms of appearance and geometry. However, obvious disagreement also can be observed, e.g., the tails of the sharks are grouped quite differently, as the same to the lips. This is due to the different ways of drawing one semantic stroke that are used by different sketches. And this kind of intra-category semantic stroke variations are further addressed by our iterative learning scheme introduced in Section~\ref{sec:iter}.

\subsection{Model learning}
DSM learning is now based on the semantic strokes output by the perceptual grouping step. Putting the semantic strokes from all training sketches into one pool (we use the sketches of mirrored pose to increase the training sketch number and flip them to the same direction), we use spectral clustering (\cite{conf/nips/Zelnik-ManorP04}) to form category-level semantic stroke clusters. Semantic strokes in one cluster possess common appearance and geometry characteristics.
Subsequently, unlike the conventional pictorial structure/deformable part-based model approach of learning parameters by optimizing on images, we follow  contour model methods by learning  model parameters from semantic stroke clusters.

\subsubsection{Spectral clustering on semantic strokes}

The clustering step forms semantic strokes into semantic stroke clusters, which will be the basic elements of the DSM. We employ spectral clustering, since it takes an arbitrary pairwise affinity matrix as input. Exploiting this, we define our own affinity measure $A_{ij}$ for semantic strokes $s_i,s_j$ whose geometrical centers are $l_i,l_j$ as $A_{ij}=exp(\frac{-K(s_i,s_j)\cdot \lVert l_i-l_j \rVert}{\rho_{s_i}\rho_{s_j}})$,
where $K(\cdot)$ is the shape context matching cost and $\rho_{s_i}$ is the local scale at each stroke $s_i$ (\cite{conf/nips/Zelnik-ManorP04}).

\begin{figure}
\centering
\includegraphics[scale=0.24]{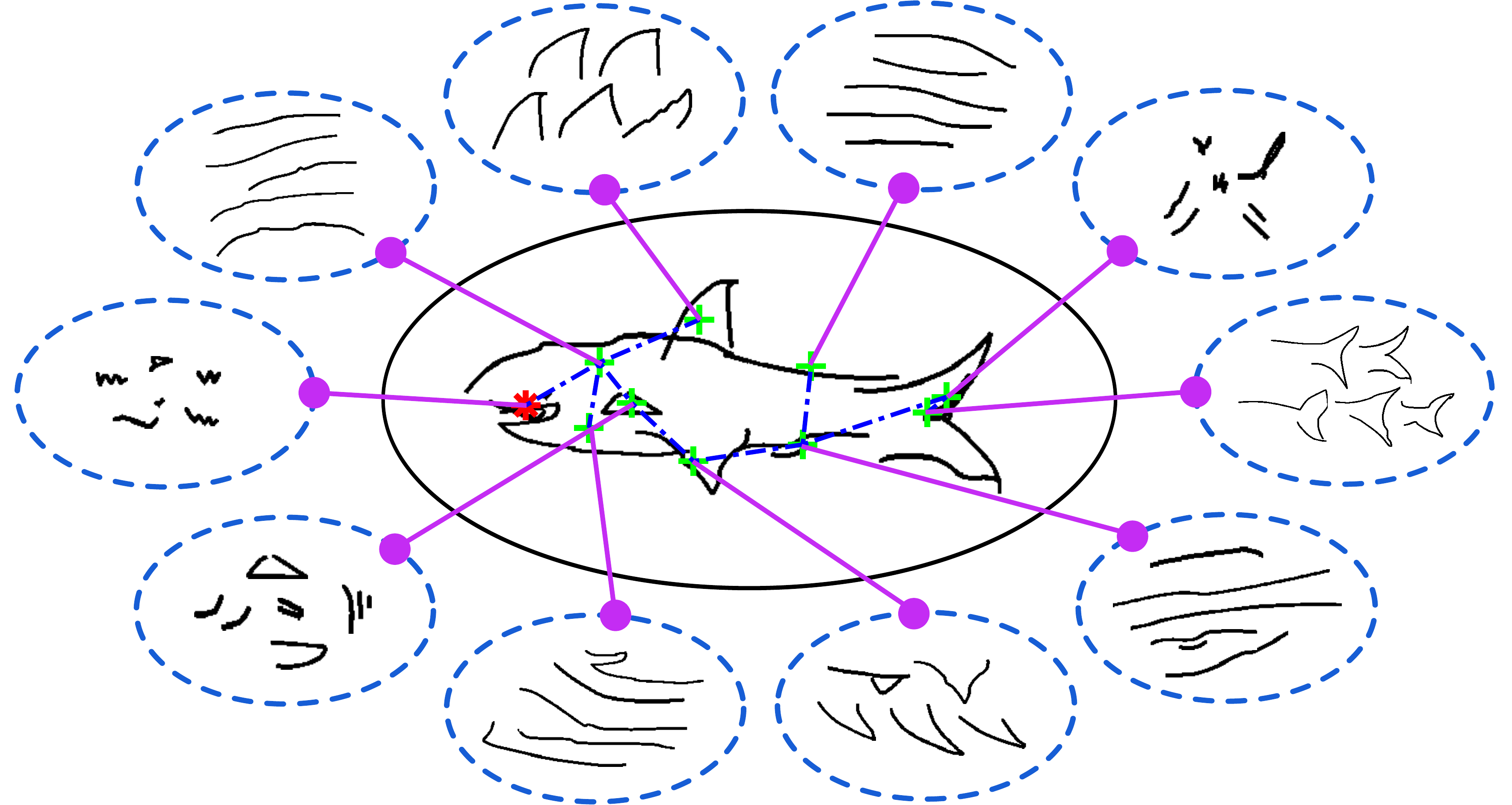}
\caption{\label{fig:elem}An example of \emph{shark} deformable stroke model with demonstration of the part exemplars in all the semantic part clusters (the blue dashed ovals), and the minimum spanning tree structure (the green crosses for tree nodes and the dash-dot lines for tree edges).} 
\end{figure}

The number of clusters discovered for each category is decided by the mean number of semantic strokes obtained by the perceptual grouper in each sketch.  
After spectral clustering, in each cluster, the semantic strokes generally agree on the appearance and location. Some cluster examples can be seen in Figure~\ref{fig:elem}.

\subsubsection{Model parameter learning}
When the semantic stroke clusters are obtained, we need to obtain the parameters $\theta$ of the model (exemplars $u$, connectivity $E$ and spatial relations $c$) to form the stroke clusters into a functional DSM. 

\noindent\textbf{Stroke exemplars}\quad We choose the $m$ strokes with the lowest average shape context matching cost to the others in each cluster $v_i$ as the stroke exemplars $u_i=\{s^a_i\}_{a=1}^{m_i}$ (\cite{Shotton:2008:MCO:1383053.1383268}). The exemplar number $m_i$ is set to a fraction of the overall stroke number in the obtained semantic stroke cluster $v_i$ according to the quality of the training data, i.e., the better the quality, the bigger the fraction. Besides, we augment the stroke exemplars with their rotation variations to achieve more precise fitting. Some learned exemplar strokes of the \emph{shark} category are shown in Figure~\ref{fig:elem}.

\noindent\textbf{Spatial Parameters}\quad 
Following the pictorial structure framework (\cite{Felzenszwalb:2005:PSO:1024426.1024429}), we treat the spatial parameters ($E$ and $c$) learning as a maximum likelihood estimation (MLE) problem and assume $E$ forms a minimum spanning tree (MST) structure. However we optimize the parameters on semantic stroke clusters rather than training images. Letting $L_i=\{l^a_i\}_{a=1}^{m_i}$ be the locations of $m_i$ strokes for cluster $v_i$ and $p(L_1,...,L_n|E,c)$ be the probability of the obtained stroke clusters' locations given the model parameters, we get:
\begin{equation}
E^*,c^*=\arg\max_{E,c}p(L_1,...,L_n|E,c).
\end{equation}
As $E$ is assumed to be a tree structure, the probability can be factorized by $E$: 
\begin{equation}
\label{equ:mle}
p(L_1,...,L_n|E,c)=\prod_{(L_i,L_j)\in E}p(L_i,L_j|c_{ij}),
\end{equation}
\begin{equation}
p(L_i,L_j|c_{ij})= \prod_{k=1}^{m_{ij}}p(l_i^k,l_j^k|c_{ij}),
\end{equation}
where $k$ indexes such stroke pairs that one stroke is from cluster $v_i$ and the other from cluster $v_j$ and they are from the same sketch. 
It can be seen that the spatial relations ${c_{ij}}$ between two clusters are independent to the edge structure $E$.
Then, we can solve this MLE problem by the following 2 steps.

\noindent\textbf{Learning the Graph Structure}\quad
To learn such a MST structure for $E$, we first need to calculate the weights of all the possible connections/edges between the clusters (the smaller the weight, the more closely correlated). 
We define edge $(v_i,v_j)$'s weight as:
\begin{equation}
w(v_i,v_j)=\prod^{m_{ij}}_{k=1}\frac{\lVert l_i^k-l_j^k\rVert}{max(height,width)}.
\end{equation}
where $height,width$ are the average dimensions of sketches. This metric ensures that stroke clusters are connected to nearby clusters, making the local spatial relations well encoded. Now we can determine the MST edge structure by minimizing 
\begin{equation}
E^*=\arg\min_E\sum_{(v_i,v_j)\in E}w(v_i,v_j).
\end{equation}
which is solved by Kruskal's algorithm. And the obtained MST is a tree that connects all the vertices and has minimum edge weights.

\noindent\textbf{Spatial relations}\quad
After the MST is learned, we can learn the spatial relations of the connected clusters. To obtain relative location parameter $c_{ij}$ for a given edge, we assume that offsets are normally distributed $p(l_i^k,l_j^k|c_{ij})=\mathcal{N}(l_i^k-l_j^k|\mu_{ij},\Sigma_{ij})$.
Then MLE result of:
\begin{equation}
(\mu_{ij}^*,\Sigma_{ij}^*)=\arg\max_{\mu_{ij}^*,\Sigma_{ij}^*}\prod^{m_{ij}}_{k=1}\mathcal{N}(l_i^k-l_j^k|\mu_{ij},\Sigma_{ij}),
\end{equation} 
straightforwardly obtains parameters $c^*_{ij} = (\mu_{ij}^*,\Sigma_{ij}^*)$. 

The learned model and edge structure is illustrated in Figures \ref{fig:teaser} and \ref{fig:elem}.



\subsection{Model matching}
As discussed in \cite{Felzenszwalb:2005:PSO:1024426.1024429}, matching
DSM to sketches or images should include two steps: model
configuration sampling and configuration energy minimization. Here, we
employ fast directional chamfer matching (FDCM,\cite{liu_cvpr2010}) as
the basic  operation of stroke registration for these two steps, which
is proved both efficient and robust at edge/stroke template matching
(\cite{conf/cvpr/ThayananthanSTC03}). In our framework, automatic
sketch model matching is used in both iterative model
training and image-sketch synthesis. This section explains this
process.

\subsubsection{Configuration sampling}
A configuration of the model $F=\{(s_i,l_i)\}_{i=1}^n$ is a model instance registered on an image.
In one configuration, exactly one stroke exemplar $s_i$ is selected in each cluster and placed at location $l_i$. Later, the configuration will be optimized by energy minimization to achieve best balance between (edge map) appearance and (model prior) geometry. Multiple configurations can be sampled, among which the best fitting can be chosen after energy minimization.

To achieve this, on a given image $I$ and for the cluster $v_i$, we first sample possible locations for all the stroke exemplars $\{s_i^a\}_{a=1}^{m_i}$ with FDCM (one stroke exemplar may have multiple possible positions). A sampling region is set based on $v_i$'s average bounding box to increase efficiency, and only positions within this region will be returned by FDCM. All the obtained stroke exemplars and corresponding locations form a set $H_m(v_i)=\{(s_i^z,l_i^z)\}_{z=1}^{h_i}(h_i \geq m_i)$. For each $(s_i^z,l_i^z)$, a chamfer matching cost $D_{cham}(s^z_i,l^z_i,I)$ will also be returned, and only the matchings with a cost under a predefined threshold will be considered by us. 

The posterior probability of a configuration $F$, according to the Bayes's rule, can be formed as:
\begin{equation}
p(F|I,\theta)\propto p(I|F,\theta)p(F|\theta),
\label{equ:postOfConf}
\end{equation}
Expanding Equation \ref{equ:postOfConf} on a stroke exemplar basis, we obtain:
\begin{equation}
p(F|I,\theta)\propto \prod_{i=1}^np(I|s_i,l_i)\prod_{(v_i,v_j)\in E}p(l_i,l_j|c_{ij}),
\end{equation}
where $p(I|s_i,l_i)$ denotes the appearance fitness for a stroke exemplar and $p(l_i,l_j|c_{ij})$ denotes the spatial relation fitness of two related stroke exemplars.

As the graph $E$ forms a MST structure, each node is dependent on a parent node except the root node which is leading the whole tree. Letting $v_{r}$ denote the root node, $C_i$ denote child nodes of $v_i$, we can firstly sample the posterior probability $p(s_r,l_r|I,\theta)$ for the root, and then sample the probability $p(s_c,l_c|s_r,l_r,I,\theta)$ for its children $\{v_c|v_c\in C_r\}$ until we reach all the leaf nodes. And we can write the marginal distribution for the root as:
\begin{align}
&p(s_r,l_r|I,\theta)\propto p(I|s_r,l_r)\prod_{v_c\in C_r}S_c(l_r),\label{equ:edf}\\
&S_j(l_i)\propto \sum_{(s_j,l_j)\in H_m(v_j)}\Bigg(p(I|s_j,l_j)p(l_i,l_j|c_{ij})\prod_{v_c\in C_j}S_c(l_j) \Bigg).\label{equ:edf}
\end{align}
$p(l_i,l_j|c_{ij})$ is the learned Gaussian offset distribution and $p(I|s_i,l_i)$ is computed from the chamfer matching cost: $p(I|s_i,l_i)=\exp(-D_{cham}(s_i,l_i,I))$.

In computation, the solution for the posterior probability of a configuration $F$ is in a dynamic programming fashion. Firstly, all the $S$ functions are computed once in a bottom-up order from the leaves to the root. Secondly, following a top-down order, we select the top $f$ probabilities $p(s_r,l_r|I,\theta)$ for the root with corresponding $f$ configurations $\{(s_r^b,l_r^b)\}_{b=1}^f$ for the root. For each root configuration $(s_r^b,l_r^b)$, we then sample a configuration for its children that have the maximum posterior probability, and we continue recursively until we reach the leaves.
From this, we obtain $f$ configurations $\{F_b\}_{b=1}^f$ for the model.

\subsubsection{Energy minimization}
Energy minimization can be considered a refinement for a configuration $F$ according to both appearances and geometry correspondences of the stroke exemplars in the input image. It is solved  similarly to configuration sampling with dynamic programming. But instead working with the posterior, it works with the energy function:
\begin{equation}
\label{equ:edf}
L^*=\arg\min_L\left(\sum^n_{i=1}D_{cham}(s_i,l_i,I)+\sum_{(v_i,v_j)\in E}D_{def}(l_i,l_j)\right),
\end{equation}
where $D_{def}(l_i,l_j)=-\log p(l_i,l_j|c_{ij})$ is the deformation cost between each stroke exemplar and its parent exemplar, and $L=\{l_i\}^n_{i=1}$ are the locations for the selected stroke exemplars in $F$. The searching space for each $l_i$ is also returned by FDCM. Comparing to configuration sampling, we set a higher threshold for FDCM, and for each stroke exemplar $s_i$ in $F$, a new series of locations $\{(s_i,l_i^k)\}$ are returned by FDCM. And a new $l_i$ is then chosen from those candidate locations $\{l_i^k\}$. To make this solvable by dynamic programming, we define:
\begin{multline}
\label{equ:qlty}
Q_j(l_i)=\min_{l_j\in\{l_j^k\}}(D_{cham}(s_j,l_j,I)\\+D_{def}(l_i,l_j)+\sum_{v_c\in C_j}Q_c(l_j)),
\end{multline}

And by combining Equations~\ref{equ:edf} and \ref{equ:qlty} and exploit the MST structure again, we can formalize the energy objective function  of the root node as:
\begin{equation}
\label{equ:rootMin}
\l^*_r=\arg\min_{l_r\in\{l^k_r\}}\left(D_{cham}(s_r,l_r,I)+\sum_{v_c\in C_r}Q_c(l_j)\right).
\end{equation} 
Through the same bottom-up routine to calculate all the $Q$ functions and the same top-down routine to find the best locations from the root to the leaves, we can find the best locations $L^*$ for all the exemplars. As mentioned before, we sampled multiple configurations and each will have a cost  after  energy minimization. We  choose the one with lowest cost as our final detection result.

\begin{figure}
\centering
\includegraphics[scale=0.145]{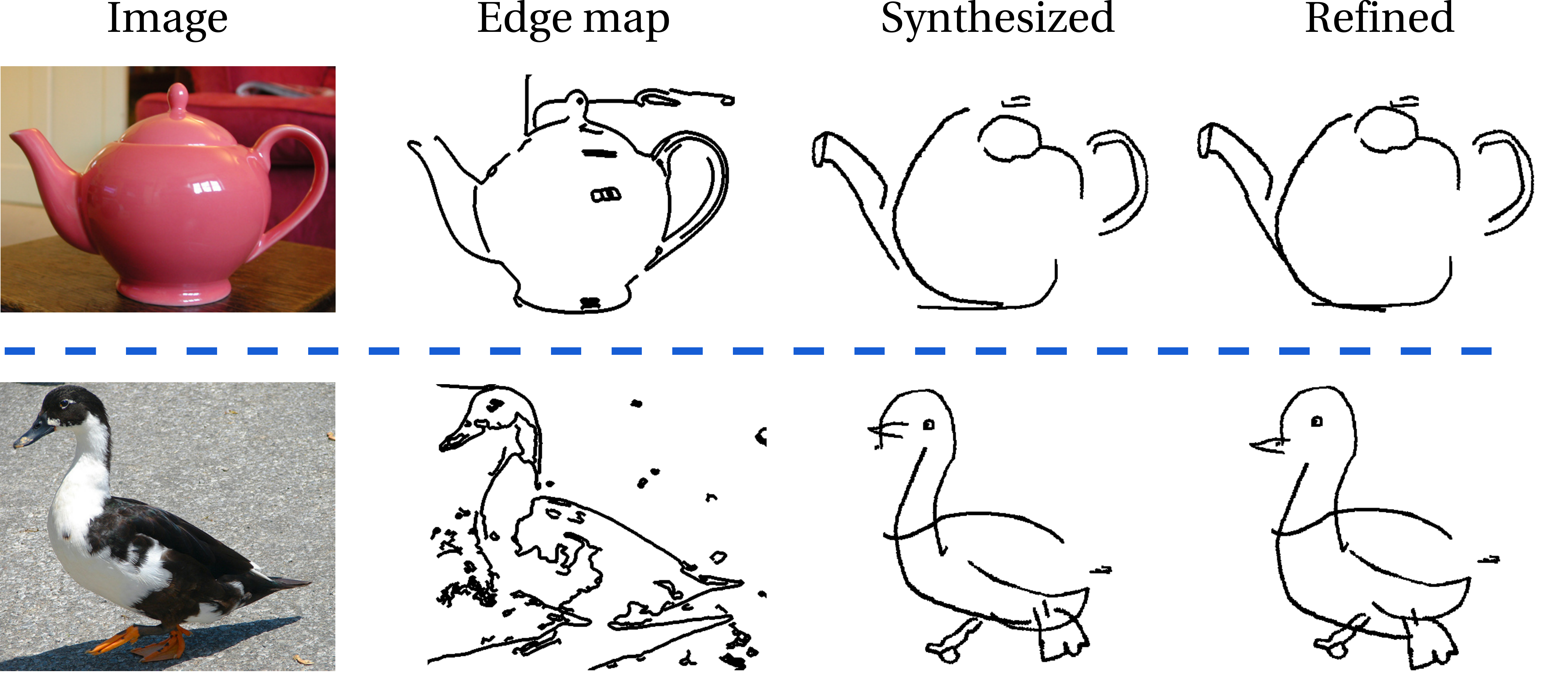}
\caption{\label{fig:refine}Refinement results illustration.}
\end{figure}

\textbf{Aesthetic refinement}\quad
The obtained detection results sometimes will have unreasonable placement for the stroke exemplar due to the edge noise. To correct this kind of error, we perform another round of energy minimization, with appearance terms 
$D_{cham}$ switched off, and rather than use chamfer matching to select the locations, we let the stroke exemplar to shift around its detection position within a quite small region.
Some refinement results are shown for the image-sketch synthesis process in Figure~\ref{fig:refine}.


\begin{figure*}[tb]
\centering
\includegraphics[scale=0.26]
{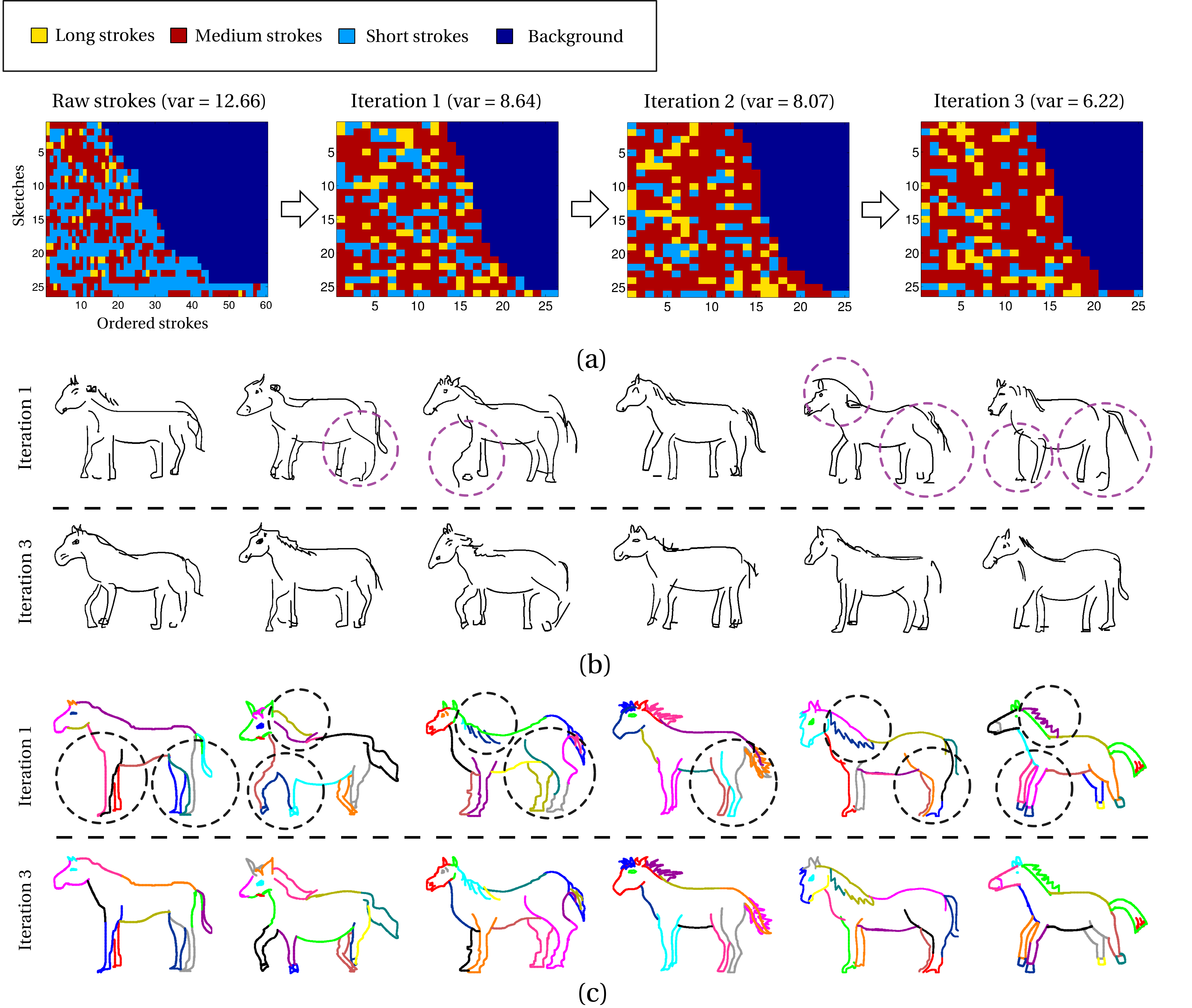}
\caption{\label{fig:it}The convergence process during model training (horse category): (a) Semantic stroke number converging process (\emph{var} denotes variance); (b) Learned horse models at iteration 1 and 3 (We pick one stroke exemplar from every stroke cluster each time to construct a horse model instance, totally 6 stroke exemplars being chosen and resulting 6 horse model instances); (c) Perceptual grouping results at iteration 1 and 3. 
Comparing to iteration 1, a much better consensus on the legs and the neck of the horse is observed on iteration 3 (flaws in iteration 1 are highlighted with dashed circles). And this is due to the increased quality of the model of iteration 3, especially on the legs and the neck parts.}
\end{figure*}

\subsection{Iterative learning}
\label{sec:iter}
As stated before, the model learned with one pass through the described pipeline is not satisfactory -- with duplicated and missing semantic strokes. To improve the quality of the model, we introduce an iterative process of: 1) perceptual grouping, 2) model learning and 3) model matching on training data in turns. The learned model will assign cluster labels for raw strokes during detection according to which stroke exemplar the raw stroke overlaps the most with or has the closest distance to. And the model labels are used in the perceptual grouping in the next iteration (Equation~\ref{eq:modelLabel}). If an overly-long stroke crosses several stroke exemplars, it will be cut into several strokes to fit the corresponding stroke exemplars.

We employ the variance of  semantic stroke numbers at each iteration as convergence metric. Over iterations, the variance decreases gradually, and we choose the semantic strokes from the iteration with the smallest variance to train the final DSM. Figure~\ref{fig:it}(a) demonstrates the convergence process of the semantic stroke numbers during the model training. Different from Figure~\ref{fig:temp}, we use 3 colors here to represent the short strokes (cyan), medium strokes (red) and long strokes (yellow). As can be seen in the figure, accompanying the convergence of stroke number variance, strokes are formed into medium strokes with properer semantics as well. Figure~\ref{fig:it}(b) illustrates the evolution of the stroke model during the training, and Figure~\ref{fig:it}(c) shows the evolution of the perceptual grouping results.

\subsection{Image-sketch synthesis}
After the final DSM is obtained from the iterative learning, it can directly be used for image-sketch synthesis through model matching on an image edge map -- where we avoid the localization challenge by assuming an approximate object bounding box has been given. Also the correct DSM (category) has to be selected in advance. And these are quite easy to be engineered in practice.

\section{Experiments}
We evaluate our sketch synthesis framework (i) qualitatively by way of showing synthesized results, and (ii) quantitatively via two user studies. We show that our system is able to generate output resembling the input image in plausible free-hand sketch style; and that it works for a number of object categories exhibiting diverse appearance and structural variations. 

We conduct experiments on 2 different datasets: (i) TU-Berlin, and (ii) Disney portrait. TU-Berlin dataset is composed of non-expert sketches while Disney portrait dataset is drawn by selected professionals. 10 testing images of each category are obtained from ImageNet, except the face category where we follow \cite{Berger:2013:SAP:2461912.2461964} to use the Center for Vital Longevity Face Database (\cite{Minear2004}). To fully use the training data of the Disney portrait dataset, we did not synthesize face category using images corresponding to training sketches of Disney portrait dataset, but instead selected 10 new testing images to synthesize from. And we normalized the grayscale range of the original sketches to 0 to 1 for the sake of simplifying the model learning process. Specifically, we chose 6 diverse categories from TU-Berlin: \emph{horse}, \emph{shark}, \emph{duck}, \emph{bicycle}, \emph{teapot} and \emph{face}; and the \emph{$90$s} and \emph{$30$s} abstraction level sketches from \emph{artist A} and \emph{artist E} from Disney portrait (\emph{$270$} level is excluded considering the high computational cost and \emph{$15$s} level is due to the presence of many incomplete sketches). 


\begin{figure*}[tb]
\centering
\includegraphics[scale=0.18]{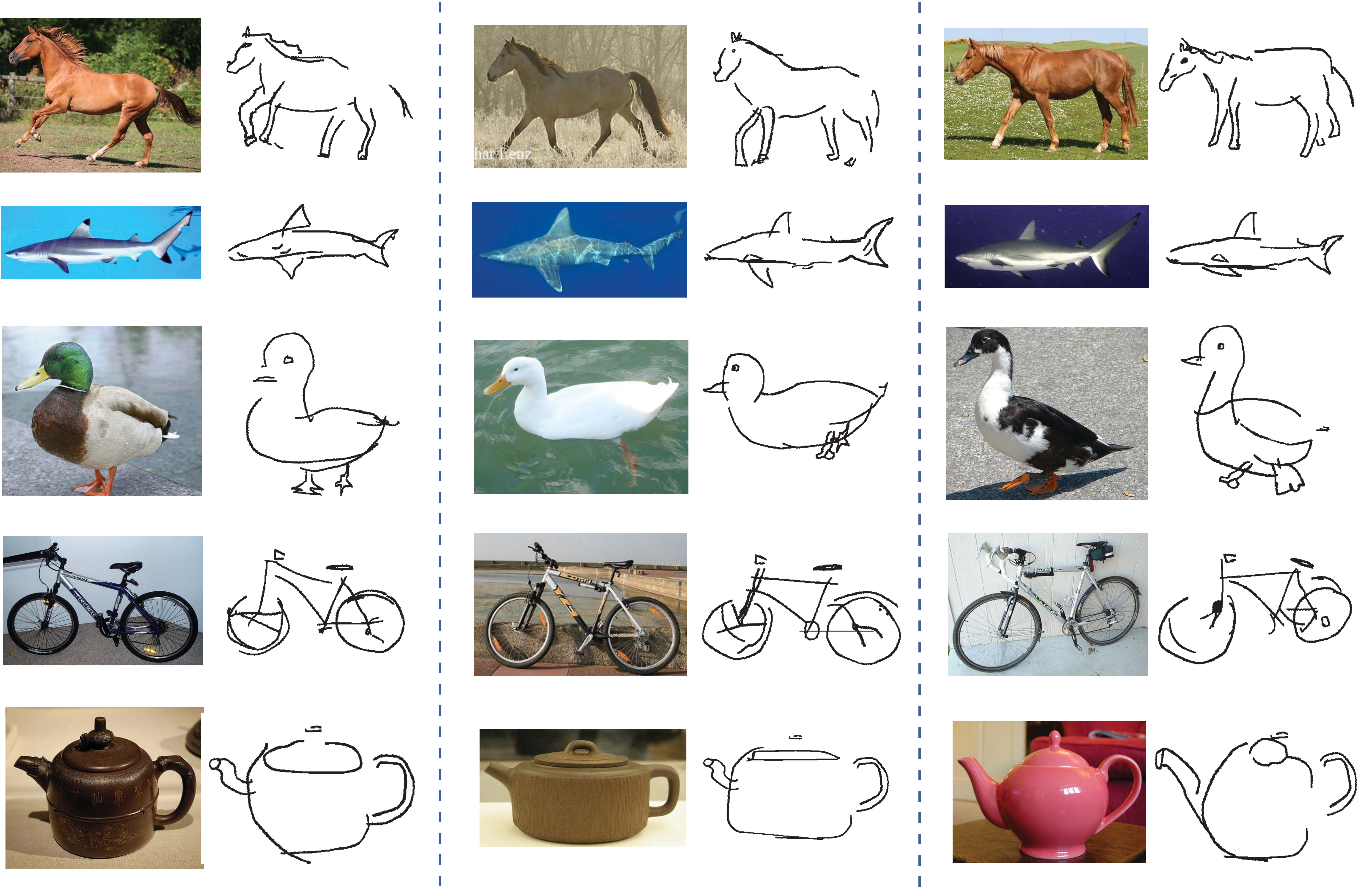}
\caption{\label{fig:synsa}Sketch synthesis results of five categories in the TU-Berlin dataset.}
\end{figure*}

\begin{figure*}[tb]
\centering
\includegraphics[scale=0.18]{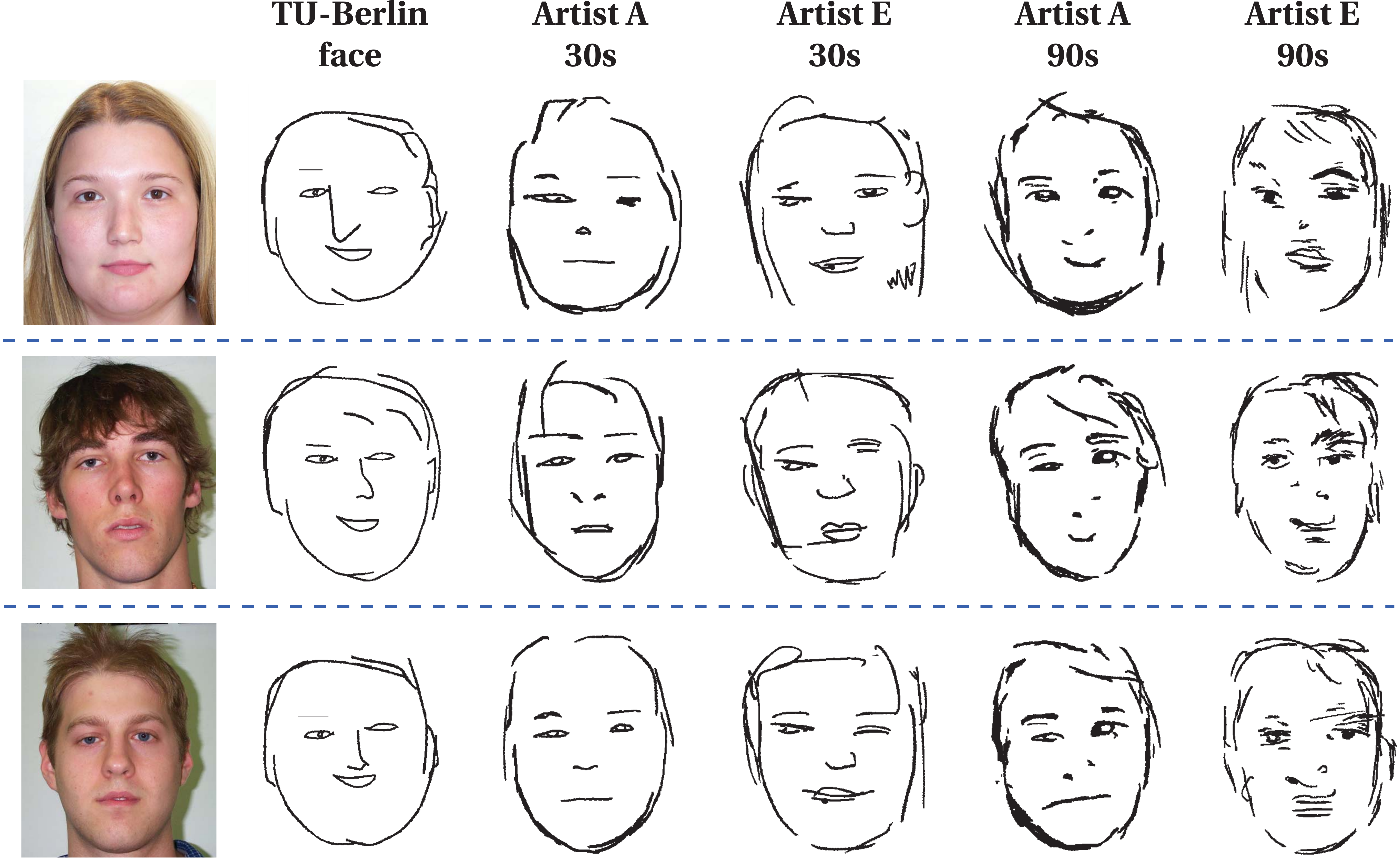}
\caption{\label{fig:synsb}A comparison of sketch synthesis results of \emph{face} category using the TU-Berlin dataset and Disney portrait dataset}
\end{figure*}

\subsection{Free-hand sketch synthesis evaluation\label{sec:Synth}}
In Figure~\ref{fig:synsa}, we illustrate synthesis results for five categories using models trained on the TU-Berlin dataset. We can see that synthesized sketches are clearly of free-hand style and abstraction while possessing  good resemblance to the input images. In particular, (i) major semantic strokes are respected in all synthesized sketches, i.e., no missing or duplicated major semantic strokes, (ii) changes in intra-category body configurations are accounted for, e.g., different leg configurations of horses, and (iii) part differences of individual objects are successfully synthesized, e.g., different styles of feet for duck and different body curves of teapots.

Figure~\ref{fig:synsb} offers synthesis results for \emph{face} only, with a comparison between these trained on the TU-Berlin dataset and Disney portrait dataset. In addition to the above observations, it can be seen that when professional datasets (e.g., portrait sketches) are used, synthesized faces tend to be more precise and resemble better the input photo. Furthermore, when compared with \cite{Berger:2013:SAP:2461912.2461964}, we can see that although without intense supervision (the fitting of a face-specific mesh model), our model still depicts major facial components with decent precision and plausibility (except for hair which is too diverse to model well), and yields similar synthesized results especially towards more abstract levels (Please refer to \cite{Berger:2013:SAP:2461912.2461964} for result comparison). We fully acknowledge that the focus of  \cite{Berger:2013:SAP:2461912.2461964} is different as compared to ours, and believe adapting detailed category-specific model alignment supervision could further improve the aesthetic quality of our results, especially towards the less abstract levels.

\subsection{Perceptual study}\label{sec:user}

Three separate user studies were performed to quantitatively evaluate our synthesis results. We employed 10 different participants for each perceptual study (to avoid prior knowledge), making a total of 20. The first user study is on sketch recognition, in which humans are asked to recognize synthesized sketches. This study confirms that our synthesized sketches are semantic enough to be recognizable by human. The second one is on perceptual similarity rating, where human subjects are asked to link the synthesized sketches to their corresponding images. By doing this, we demonstrate the intra-category discrimination power of our synthesized sketches. 

\noindent\textbf{Sketch recognition}\quad Sketches synthesized using models trained on TU-Berlin dataset are used in this study, so that human recognition performance reported in \cite{eitz2012hdhso} can be used as comparison. There are 60 synthesized sketches in total, with 10 per category. We equally assign 6 sketches (one from each category) to every participant and ask them to select an object category for each sketch (250 categories are provided in a similar scheme as in \cite{eitz2012hdhso}, thus chance is 0.4\%). From Table \ref{tab:recog}, we can observe that our synthesized sketches can be clearly recognized by humans, in some cases offering 100\% accuracy. It can be further noted that human recognition performance on our sketches follows a very similar trend across categories to that reported in \cite{eitz2012hdhso}. The overall higher performance of ours is most likely due to the much smaller scale of our study. The result of this study clearly shows that our synthesized sketches convey enough semantic meaning and are highly recognizable as human-drawn sketches.

\begin{table}[htbp]
\centering
\scriptsize
\caption{\label{tab:recog}Recognition rate of human users for (S)ynthesised and (R)eal  sketches \protect(\cite{eitz2012hdhso}).}
\begin{tabular}{ccccccc}
\toprule
 & \textbf{Horse} & \textbf{Shark}  & \textbf{Duck}  &\textbf{Bicycle} & \textbf{Teapot}  &\textbf{Face} \\ \hline
\textbf{S}  & 100\%& 40\%&100\%&100\%&90\%&80\% \\ \hline
\textbf{R}  & 86.25\%& 60\%&78.75\%&95\%&88.75\%&73.75\% \\ \bottomrule
\end{tabular}
\end{table}

\noindent\textbf{Image-sketch similarity}\quad For the second study, both TU-Berlin dataset and Disney portrait dataset are used. In addition to the 6 models from TU-Berlin, we also included 4 models learned using the \emph{$90$s} and \emph{$30$s} level sketches from \emph{artist A} and \emph{artist E} from Disney portrait dataset. For each category, we randomly chose 3 image pairs, making 30 pairs (3 pairs $\times$ 10 categories) in total for each participant. Each time, we show the participant one pair of images and their corresponding synthesized sketches, where the order of sketches may be the same or reversed as the image order (Due to the high abstraction nature of the sketches, only a pair of sketch is used and two corresponding images are provided for clues each time). Please refer to Figure \ref{fig:synsa} to see some example image and sketch pairs. The participant is then asked to decide if the sketches are of the same order as the images. We consider a choice to be correct if the participant correctly identified the right ordering. Finally, the accuracy for each category is averaged over 30 pairs and summarized in Table \ref{tab:link}. A binomial test is applied to the results, and we can see that, except \emph{duck} and \emph{Artist E 90s}, all the rest results are significantly better than random guess (50\%), with most $p<0.01$.  The relatively weaker performance for \emph{duck} and \emph{teapot} from TU-Berlin is mainly due to a lack of training sketch variations as opposed to image domain, resulting in the model failing to capture enough appearance variations in images. On Disney portrait dataset, matching accuracy is generally on the same level as TU-Berlin, yet there appears to be a big divide on \emph{artist E 90s}. This is self-explanatory when one compares synthesized sketches of the 90s level from \emph{artist E} (last column of Figure~\ref{fig:synsb}) with other columns -- \emph{artist E 90s} seems to depict a lot more short and detailed strokes making the final result relatively messy. In total, we can see that our synthesized sketches possess sufficient intra-category discrimination power.

\begin{table}[htbp]
\centering
\scriptsize
\caption{\label{tab:link} Image-sketch similarity rating experiment results.}
\begin{tabular}{cccccc}
\toprule
\multicolumn{1}{}{} & \textbf{Horse} & \textbf{Shark} & \textbf{Duck} & \textbf{Bicycle} & \textbf{Teapot} \\ \hline
\textbf{Acc.} & 86.67\% & 73.33\% & 63.33\% & 83.33\% & 66.67\% \\ \hline
\textbf{p} & $<0.01$ & $<0.01$ & 0.10 & $<0.01$ & $<0.05$ \\ \hline
\textbf{} & \textbf{Face} & \textbf{A 30s} & \textbf{E 30s} & \textbf{A 90s} & \textbf{E 90s} \\ \hline
\textbf{Acc.} & 76.67\% & 76.67\% & 90.00\% & 73.33\% & 56.67\% \\ \hline
\textbf{p} & $<0.01$ & $<0.01$ & $<0.01$ & $<0.01$ & 0.29 \\ 
\bottomrule
\end{tabular}
\end{table}

\subsection{Parameter tuning}
Our model is intuitive to tune, with important parameters constrained within perceptual grouping. There are two sets of parameters affecting model quality: semantic stroke length and weights for different terms in Equation~\ref{equ:pegr}. Semantic stroke length reflects negatively to the semantic stroke number and it needs to be tuned consistent with the statistical observation of that category. And it is  estimated as the $\lambda$ illustrated in the \emph{stroke length} term in Section~\ref{sec:percepGroup}. For $\eta_{sem}$ we used $1$-$3$ for TU-Berlin dataset and the \emph{$30$s} level portrait sketches, and for the \emph{$90$s} level portrait sketches, $\eta_{sem}$ is set 8 and 11 respectively for the \emph{90s} level of \emph{artist A} and \emph{artist E}. This is because in the less abstracted sketches artists tend to use more short strokes to form one semantic stroke. For those categories with $\eta_{sem}=1$, we found 85\%-95\% of the maximum stroke length is a good range to tune against for $\tau$ since our earlier stroke-level study suggests semantic stroke strokes tend to cluster within this range (see Figure~\ref{fig:strlen1}). 

Regarding weights for different terms in Equation \ref{equ:pegr}, we used the same parameters for both the TU-Berlin dataset and \emph{$30$s} level portrait sketches, and set $\omega_{pro}$, $\omega_{con}$ and $\omega_{len}$ (for proximity, continuity and stroke length respectively) uniformly to $0.33$. For the \emph{$90$s} level sketches, again since too many short strokes are used, we switched off the continuity term, and set $\omega_{pro}$ and $\omega_{len}$ both to $0.5$. The weight $\omega_{sim}$ and adjustment factors $\mu_{temp}$ and $\mu_{mod}$ (corresponding to similarity, local temporal order and model label) are all fixed as $0.33$ in all the experiments.

%

\section{Further discussions}
\noindent\textbf{Data alignment:}\quad  Although our model can address a good amount of variations in the number, appearance and location of parts without the need for well-aligned datasets, a poor model may be learned if the topology diversity (existence, number and layout of parts) of the training sketches is too extreme. This could be alleviated by selecting fine-grained sub-categories of sketches to train on, which would require more constrained collection of training sketches.



\noindent\textbf{Model quality:}\quad Due to the unsupervised nature of our model, it has difficulty modelling  challenging objects with complex inner structure. For example, buses often exhibit complicated features such as the number and location of windows. We expect that some simple user interaction akin to that used in interactive image segmentation might help to increase model precision, for example by asking the user to scribble an outline to indicate rough object parts.

Another weakness of our model is that the diversity of synthesized results is highly dependent on training data. If there are no similar sketches in the training data that can roughly resemble the input image, it will be hard to generate a good looking free-hand sketch for that image, e.g., some special shaped teapot images. We also share the common drawback of part-based models, that severe noise will affect detection accuracy.

\noindent\textbf{Aesthetic quality:}\quad  In essence, our model learns a normalized representation for a given category. However, apart from common semantic strokes, some individual sketches will exhibit unique parts not shared by others, e.g., saddle of a horse. To explicitly model those accessory parts can significantly increase the descriptive power of the stroke model, and thus is an interesting direction to explore in the future.  Last but not least, as the main aim of this work is to tackle the modeling for  category-agnostic sketch synthesis, only very basic aesthetic refinement post-processing was employed. A direct extension of current work will be therefore leveraging advanced rendering techniques from the NPR domain to further enhance the aesthetic quality of our synthesized sketches.

\section{Conclusion}

We presented a free-hand sketch synthesis system that for the first time works outside of just one object category. Our model is data-driven and uses publicly available sketch datasets regardless of whether drawn by non-experts or professionals. With minimum supervision, i.e., the user selects a few sketches of similar poses from one category, our model automatically discovers common semantic parts of that category, as well as encoding structural and appearance variations of those parts. Importantly, corresponding pairs of photo and sketch images are not required for training, nor any alignment is required. By fitting our model to an input image, we automatically generate a free-hand sketch that shares close resemblance to that image. Results provided in the previous section confirms the efficacy of our model.


\bibliographystyle{spbasic}      
\bibliography{template}   


\end{document}